%% file: main.tex
\documentclass[runningheads]{llncs}

\usepackage{eccv}

\usepackage{eccvabbrv}

\usepackage{graphicx}
\usepackage{booktabs}

\usepackage[accsupp]{axessibility}  

\usepackage{xcolor}
\definecolor{myblue}{RGB}{0,55,119}
\definecolor{myred}{RGB}{177,0,28}
\definecolor{myorange}{RGB}{253,102,31}
\definecolor{mygreen}{RGB}{79,143,0}
\definecolor{mypurple}{RGB}{129,52,190}
\definecolor{mygrey}{RGB}{102, 102, 102}
\definecolor{mybrightblue}{RGB}{0,150,255}

\usepackage[hidelinks,pagebackref]{hyperref}

\usepackage{orcidlink}

\hypersetup{colorlinks=true}

\begin{document}

\title{SPEAR: A Simulator for\\Photorealistic Embodied AI Research}

\titlerunning{SPEAR: A Simulator for\\Photorealistic Embodied AI Research}

\author{
Mike Roberts\inst{1,2} \and
Renhan Wang\inst{3} \and
Rushikesh Zawar\inst{4} \and
Rachith Dey-Prakash\inst{2} \and \\
Quentin Leboutet\inst{2} \and
Stephan R.~Richter\inst{2} \and
Matthias M{\"u}ller\inst{2} \and
German Ros\inst{5} \and \\
Rui Tang\inst{3} \and
Stefan Leutenegger\inst{6,7} \and
Yannick Hold-Geoffroy\inst{1} \and
Kalyan Sunkavalli\inst{1} \and
Vladlen Koltun\inst{2}}

\authorrunning{M.~Roberts et al.}

\institute{$^1$Adobe Research ~ $^2$Intel Labs ~ $^3$Manycore Tech Inc ~ $^4$Adobe ~ $^5$NVIDIA \\
$^6$ETH Zurich ~ $^7$Imperial College London\\
\vspace{9pt}\url{https://github.com/spear-sim/spear}\vspace{-18pt}}

\maketitle

\input{00_abstract}
\input{01_introduction}
\input{02_related_work}
\input{03_programming_model}
\input{04_system_architecture}
\input{05_results}

\input{06_conclusions}
\input{07_acknowledgments}

\clearpage

%
%
\bibliographystyle{splncs04}
\bibliography{main}
\end{document}

%% file: 00_abstract.tex
\begin{abstract}
Interactive simulators have become powerful tools for training embodied agents and generating synthetic visual data, but existing photorealistic simulators suffer from limited generality, programmability, and rendering speed.
We address these limitations by introducing \emph{SPEAR: A Simulator for Photorealistic Embodied AI Research}.
At its core, SPEAR is a Python library that can connect to, and programmatically control, any Unreal Engine (UE) application via a modular plugin architecture.
SPEAR exposes over 14K unique UE functions to Python, representing an order-of-magnitude increase in programmable functionality over existing UE-based simulators.
Additionally, a single SPEAR instance can render 1920$\times$1080 photorealistic beauty images directly into a user's NumPy array at 73 frames per second -- an order of magnitude faster than existing UE plugins -- while also providing ground truth image modalities that are not available in any existing UE-based simulator (e.g., a non-diffuse intrinsic image decomposition, material IDs, and physically based shading parameters).
Finally, SPEAR introduces an expressive high-level programming model that enables users to specify complex graphs of UE work with arbitrary data dependencies among work items, and to execute these graphs deterministically within a single UE frame.
We demonstrate the utility of SPEAR through a diverse collection of example applications: controlling multiple embodied agents with distinct action spaces (e.g., humans, cars, and robots) across several in-the-wild UE projects; rendering photorealistic city-scale environments; manipulating UE's procedural content generation systems; rendering synchronized multi-view images of detailed human faces; coordinating an interactive co-simulation with the MuJoCo physics simulator; and editing scenes with natural language via an AI coding assistant.
\end{abstract}

%% file: 01_introduction.tex
\section{Introduction}
\label{sec:intro}

\begin{figure}[t]
  \centering
  \includegraphics[width=11.3cm]{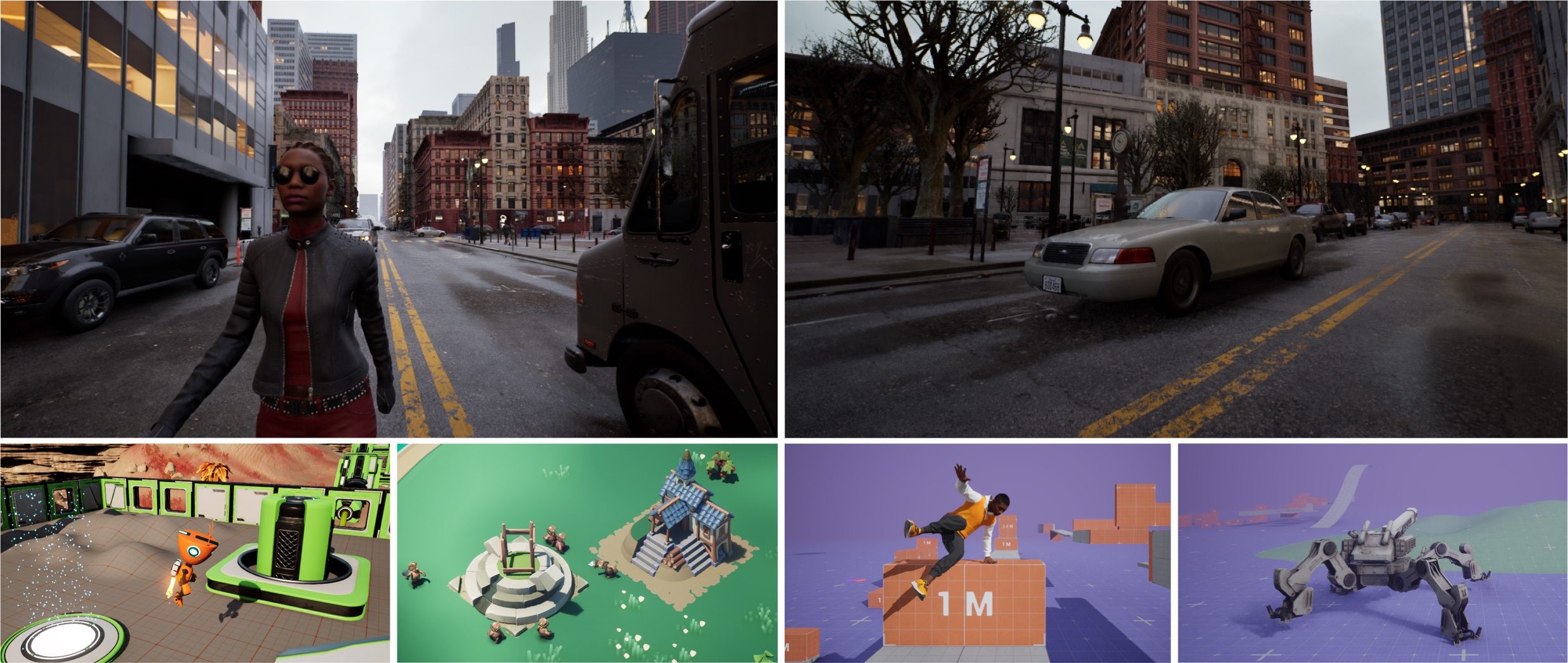}\vspace{-5pt}
  \caption{
SPEAR is a Python library that can connect to, and programmatically control, any Unreal Engine (UE) application via a modular plugin architecture.
SPEAR exposes over 14K unique UE functions, representing an order-of-magnitude increase in programmable functionality over existing simulators.
We demonstrate the flexibility of SPEAR by using it to control 6 distinct embodied agents (each with a different action space) across several Epic Games sample projects: a person and a car from \textsc{CitySample} (top); a flying robot from \textsc{StackOBot} (bottom far left); multiple agents in a resource collecting game called \textsc{CropoutSample} (bottom center left); as well as a person with parkour skills and a quadruped robot from \textsc{GameAnimationSample} (bottom right).
}
  \label{fig:control_sample_projects}\vspace{-15pt}
\end{figure}

\begin{figure}[t]
  \centering
  \includegraphics[width=11.3cm]{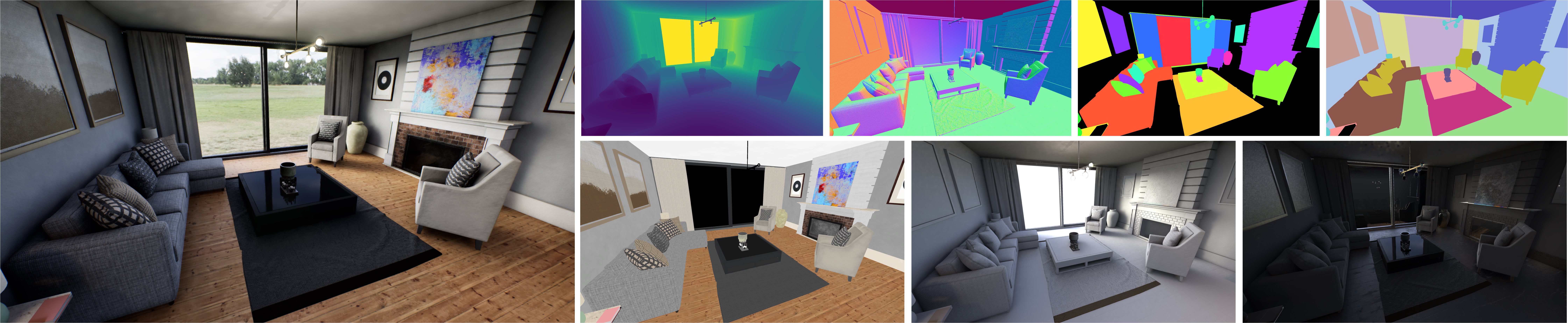}\vspace{-5pt}
  \caption{
SPEAR includes a customizable camera sensor that can render 1920$\times$1080 photorealistic beauty images (left) directly into a user's NumPy array at 73 frames per second -- an order of magnitude faster than existing UE plugins -- while also providing ground truth image modalities that are not available in any existing UE-based simulator. For example, the SPEAR camera sensor can render all of the image modalities in the Hypersim dataset~\cite{roberts:2021}, i.e., depths, surface normals, instance and semantic IDs (right top), and a non-diffuse intrinsic image decomposition (right bottom), as well as material IDs and physically based shading parameters (see our supp.~material).
}
  \label{fig:hypersim}\vspace{-15pt}
\end{figure}

\begin{figure}[t]
  \centering
  \includegraphics[width=12.0cm]{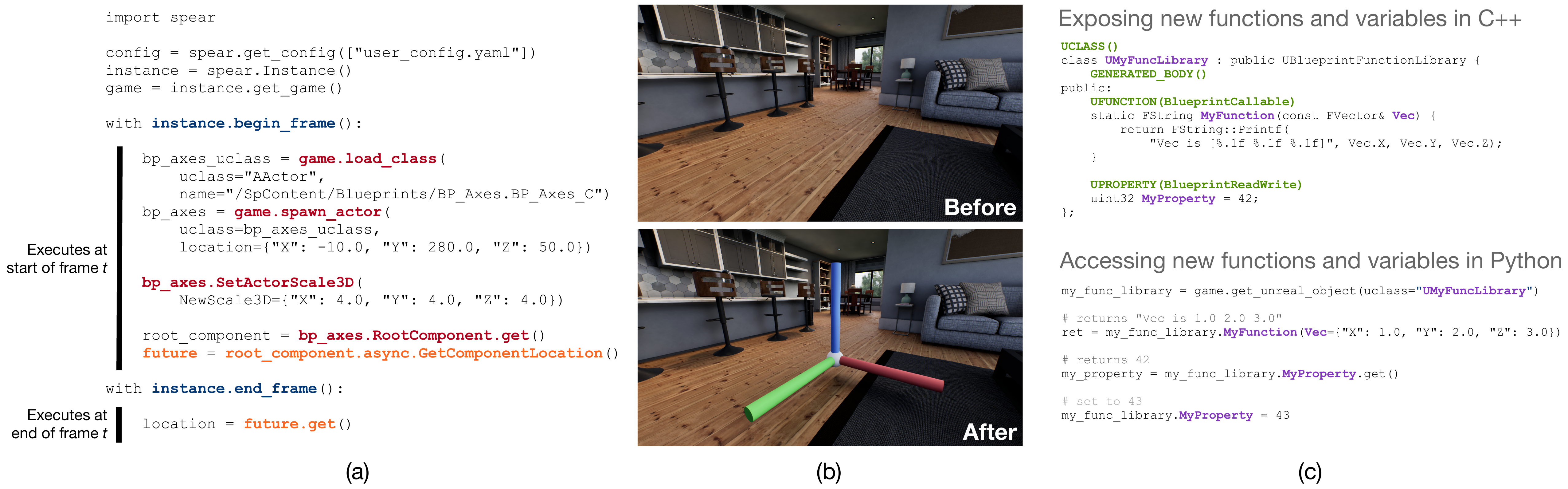}\vspace{-5pt}
  \caption{
\textbf{(a, b):} We demonstrate several fundamental concepts in the SPEAR programming model with a simple example program that spawns a set of coordinate axes in an indoor environment.
In our programming model, graphs of UE work are specified as transactions.
In particular, the user specifies a transaction by defining a \texttt{\textcolor{myblue}{begin\_frame}} context followed by an \texttt{\textcolor{myblue}{end\_frame}} context.
Within each context, the user specifies a graph of UE work simply by implementing it as Python code.
Any C++ function or variable that is visible to UE's reflection system (e.g., \texttt{\textcolor{myred}{SetActorScale3D}}, \texttt{\textcolor{myred}{RootComponent}}) can be accessed as though it was a native Python function or attribute.
For improved efficiency, we provide an asynchronous variant for each function in SPEAR (e.g., \texttt{\textcolor{myorange}{async.GetComponentLocation}}) that avoids synchronizing with UE.
\textbf{(c):} It is trivial to expose new C++ functions and variables to UE's reflection system, and therefore to SPEAR, simply by adding a \texttt{\textcolor{mygreen}{UFUNCTION}} or \texttt{\textcolor{mygreen}{UPROPERTY}} annotation next to the function or variable in any C++ header file.
}
  \label{fig:programming_model}
\end{figure}

Interactive simulators have become a critical layer of scientific infrastructure, participating in major breakthroughs in reinforcement learning (e.g., world-champion performance in competitive e-sports games~\cite{openai:2019:dota,vinyals:2017,vinyals:2019,wurman:2022} and drone racing~\cite{kaufmann:2018,kaufmann:2020,kaufmann:2023,loquercio:2020,loquercio:2021}), sensorimotor control~\cite{richter:2016,richter:2017,zhou:2018}, autonomous driving~\cite{dosovitskiy:2017,muller:2018}, dexterous manipulation~\cite{openai:2019:rubiks}, and quadruped locomotion~\cite{hwangbo:2019,lee:2020,miki:2022}.
Additionally, photorealistic synthetic datasets (e.g.,~\cite{roberts:2021}) are now being used to train and evaluate large-scale foundation models (e.g.,~\cite{bachmann:2024,gabeur:2026,mizrahi:2023,tong:2024,simeoni:2025,yang:2025}), as well as state-of-the-art methods for 3D and 4D reconstruction~\cite{mazur:2024,wang:2025,wang:2026}, segmentation~\cite{kirillov:2023,ravi:2025}, depth estimation~\cite{ke:2024}, and controllable video synthesis~\cite{bai:2025}.

Motivated by the parallel goals of interactivity and photorealism, the computer vision, robotics, and embodied AI communities are increasingly leveraging photorealistic simulators that are built on top of industrial-strength game engines like Unreal~\cite{bordes:2023,dosovitskiy:2017,lerer:2016,martinezgonzalez:2019,martinezgonzalez:2021,qiu:2017,shah:2017,unrealengine:2026,ye:2025,zhong:2025}, Unity~\cite{gan:2021,juliani:2018,kolve:2017,unity:2026}, and Panda3D~\cite{panda3d:2026,wu:2025a}, as well as closed-source commercial simulation platforms like IsaacSim~\cite{ge:2024,gong:2023,li:2022,nvidia:2026,wu:2025b}.
Among these alternatives, the Unreal Engine (UE)~\cite{unrealengine:2026} is a compelling platform due to its state-of-the-art real-time rendering~\cite{karis:2021,narkowicz:2025,wright:2022} and procedural content generation systems~\cite{langmead:2025}, its proven track record in commercial game development~\cite{epic:2025,obedkov:2025}, and because it is completely open-source.

However, existing simulators that build on UE have several important limitations.
First, all existing UE-based simulators expose relatively limited hand-crafted Python interfaces (i.e., consisting of hundreds of functions).
Second, existing UE-based simulators can incur significant communication overhead when returning large blocks of data (e.g., high-resolution images) to a user's Python program.
For example, in the UE-based simulators we tested, we found that returning a 1920$\times$1080 image to Python can be 20--35$\times$ slower than rendering the same image to the viewport in a standalone UE application with no Python communication.
Third, most UE-based simulators are distributed as large monolithic applications, in some cases requiring their own custom fork of the 10+ million line UE codebase, rather than as modular plugins.
As a result, it is not straightforward to integrate these simulators into existing projects, and likewise, it is not straightforward to integrate third-party assets into existing simulators.

In this work, we introduce SPEAR, a \textbf{s}imulator for \textbf{p}hotorealistic \textbf{e}mbodied \textbf{A}I \textbf{r}esearch that addresses all of the limitations described above (see Figures~\ref{fig:control_sample_projects} and~\ref{fig:hypersim}).
Throughout the design and implementation of SPEAR, our key technical insight is to expose a comprehensive C++ interface for interacting directly with UE's runtime reflection system~\cite{maes:1987,unrealengine:2026:reflection}, thereby enabling our Python library to find classes, call functions, and manipulate variables on objects dynamically at runtime using strings as keys (i.e., without requiring hand-crafted wrapper code for each class, function, and variable).
By hooking directly into UE's reflection system, SPEAR exposes a significant amount of underlying UE functionality (over 14K functions) while maintaining a modest code footprint (see Table~\ref{tbl:api_comparison}).

On top of our low-level reflection interface, we introduce an expressive high-level programming model that enables users to call any UE function as though it was a native Python function, provided the UE function is visible to the reflection system (see Figure~\ref{fig:programming_model}).
For improved efficiency, we provide an asynchronous variant for every function in SPEAR that avoids synchronizing with UE, and we implement a mechanism for passing NumPy arrays back and forth between UE functions and user Python code without requiring any data copying.

Together, these novel technical contributions enable an order-of-magnitude increase in programmable functionality over existing UE-based simulators, while also enabling us to render 1920$\times$1080 photorealistic beauty images directly into a user's NumPy array at 73 frames per second -- an order of magnitude faster than existing UE plugins (see Table~\ref{tbl:ablation}).

We demonstrate the utility of SPEAR through a diverse collection of example applications: controlling multiple embodied agents with distinct action spaces (e.g., humans, cars, and robots) across several in-the-wild UE projects; rendering photorealistic city-scale environments; manipulating UE's procedural content generation systems; rendering synchronized multi-view images of detailed human faces; coordinating an interactive co-simulation with the MuJoCo physics simulator~\cite{todorov:2012}; and editing scenes with natural language via an AI coding assistant.\looseness=-1



%% file: 02_related_work.tex
\section{Related Work}
\label{sec:related_work}

\noindent\textbf{Simulators for Embodied AI}~~
We discuss interactive simulators based on game engines and commercial simulation platforms in Section \ref{sec:intro}.
Additionally, the embodied AI community has developed simulators for navigating through static scans of real-world environments~\cite{savva:2017,savva:2019,xia:2018}, and manipulating fully interactive environments~\cite{li:2021,puig:2024,shen:2021,szot:2021,xiang:2020}, using custom simulation stacks that typically prioritize simulation speed over photorealism.
In contrast, we choose to build on top of UE to leverage its state-of-the-art rendering and procedural content generation systems, as well as its vibrant ecosystem of high-fidelity content.

\noindent\textbf{Using Games as AI Simulators}~~
Spanning across the history of the medium, commercial video games have been successfully repurposed into AI simulators, from the earliest Atari 2600 games~\cite{bellemare:2013,mnih:2015}, to early first-person 3D games~\cite{beattie:2016,jaderberg:2019,kempka:2016,zhou:2018}, to modern strategy~\cite{openai:2019:dota,vinyals:2017,vinyals:2019}, racing~\cite{wurman:2022}, and open-world exploration~\cite{fan:2022,johnson:2016,krahenbuhl:2018,magne:2026,richter:2016,richter:2017} games.
However, these simulators either require a hand-crafted scripting interface to be implemented by the game's developers (e.g.,~\cite{vinyals:2017,wurman:2022}); or are implemented as unintrusive wrappers that intercept graphics driver communication (e.g.,~\cite{richter:2016,richter:2017,krahenbuhl:2018}) and programmatically inject device input (e.g.,~\cite{magne:2026}), and therefore offer limited opportunities to observe and manipulate the underlying game state.
In contrast, we choose to implement our simulator as a set of modular plugins that provide comprehensive programmatic access to UE, and can be enabled in any UE application by adding a single-line declaration to a project file.

\noindent\textbf{Scripting Interfaces to the Unreal Engine}~~
UE includes a powerful node-based visual programming environment known as \emph{Blueprints}~\cite{unrealengine:2026:blueprints}, which exposes nearly every class, function, and property that is visible to the reflection system, and can be used to script standalone applications.
But Blueprints are unfamiliar to most AI practitioners, and they are stored in a proprietary binary data format that is incompatible with text-based developer workflows (e.g., diffing and merging tools, AI coding assistants).
On the other hand, the Unreal Editor includes a self-contained Python environment that exposes a comparable amount of functionality to Blueprints~\cite{unrealengine:2026:python}.
However, this Python environment is intended for automating offline content-creation tasks inside the Unreal Editor and cannot be used to script standalone applications.
In our work, we expose a similar breadth of functionality to Blueprints and the Unreal Editor, and our Python interface can be used to script standalone applications, live simulations running inside the editor, UE's path tracer, and the editor itself, all through a unified and expressive high-level programming model.

%% file: 03_programming_model.tex
\vspace{-5pt}
\section{Programming Model}
\label{sec:programming_model}
\vspace{-5pt}

\begin{figure}[t]
  \centering
  \includegraphics[width=12.0cm]{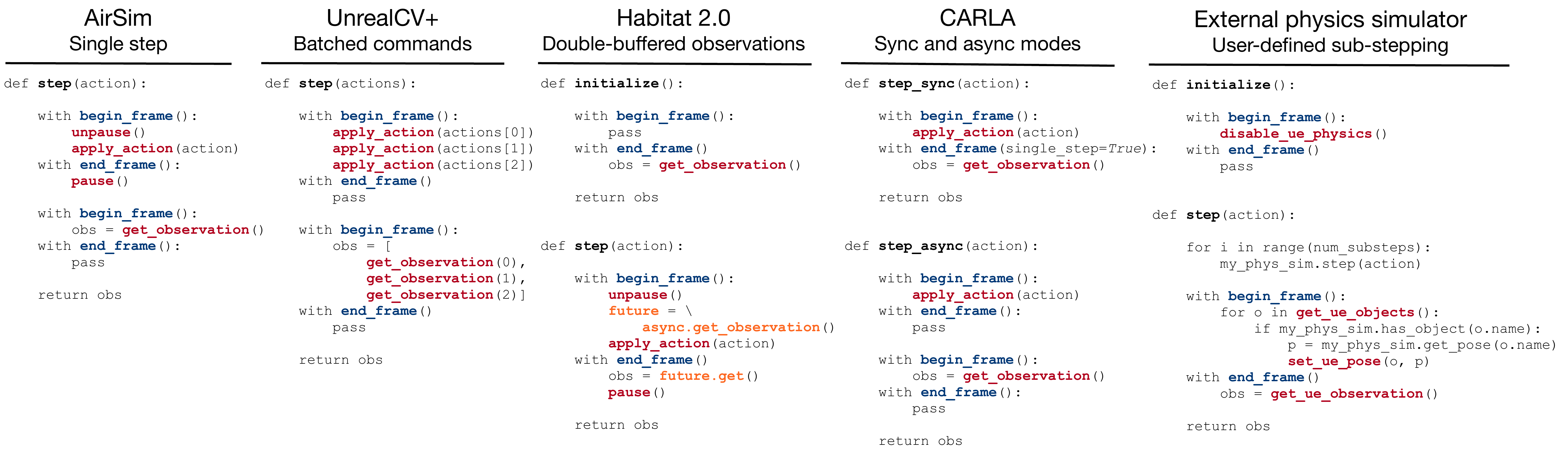}\vspace{-5pt}
  \caption{
We demonstrate the flexibility of our programming model by using it to implement a variety of synchronization strategies found in existing simulators.
For each strategy, we implement a simplified OpenAI Gym \texttt{step} function~\cite{brockman:2016}, which takes an \texttt{action} as input, steps a simulation forward, and returns an \texttt{observation} as output.
In our programming model, it is straightforward to implement AirSim's approach for single-stepping~\cite{shah:2017}, UnrealCV+'s approach for batched commands~\cite{qiu:2017,zhong:2025}, Habitat 2.0's approach for double-buffered observations~\cite{szot:2021}, CARLA's synchronous and asynchronous stepping modes~\cite{dosovitskiy:2017}, and co-simulation via an external physics simulator (e.g.,~\cite{todorov:2012}) with user-defined sub-stepping.
}
  \label{fig:step_functions}\vspace{-15pt}
\end{figure}

\begin{figure}[t]
  \centering
  \includegraphics[width=12.0cm]{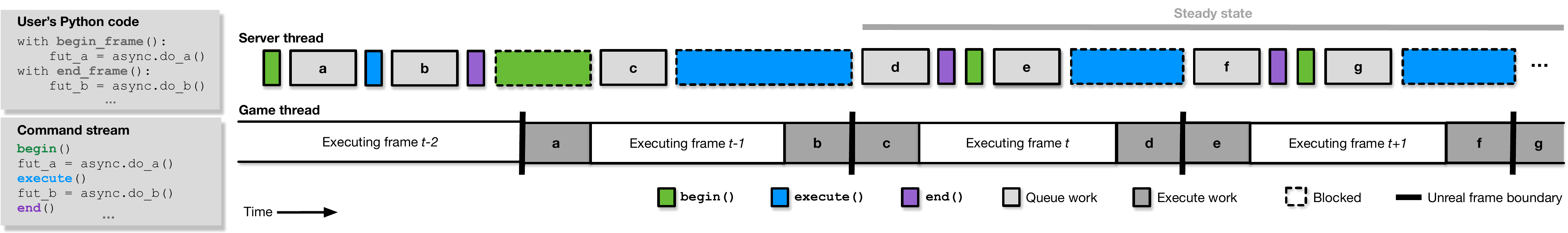}\vspace{-5pt}
  \caption{
Illustration of how the asynchronous operations in our programming model can be used to completely avoid blocking the UE game thread, thereby enabling a UE application to execute user Python code at native frame rates.
Our Python client transforms a user's Python code (top left) into a stream of commands (bottom left), and these commands are sent to our server running on its own thread within a UE application.
We assume that the user's Python code consists of only asynchronous operations, and therefore the server can avoid synchronizing with the game thread when queuing these operations (light grey boxes).
When necessary, the server intentionally blocks (dotted colored boxes) to prevent the user's Python code from getting too far ahead of the game thread.
Once the game thread reaches the beginning of frame $t$, the system is in a steady state.
In this steady state, as long as the user's \texttt{\textcolor{mygrey}{begin\_frame}} and \texttt{\textcolor{mygrey}{end\_frame}} contexts execute faster than a UE frame, i.e., as long as the width of the sequence \texttt{[~\textbf{d},~\textbf{\textcolor{mypurple}{end}},~\textbf{\textcolor{mygreen}{begin}},~\textbf{e},~\textbf{\textcolor{mybrightblue}{execute}}~]} on the server thread is less than the width of frame $t$ on the game thread in the diagram above, then the UE application will execute at its native frame rate.
}
  \label{fig:steady_state}\vspace{-15pt}
\end{figure}

In this section, we describe the design goals, fundamental concepts, and formal guarantees provided by our programming model (see Figures~\ref{fig:programming_model} and~\ref{fig:step_functions}).
In Section~\ref{sec:system_architecture}, we discuss our implementation of this model, including our low-level reflection interface and our method for passing NumPy arrays to and from UE.
In our supplementary material, we include additional details about our model, including how we can use it to control the Unreal Editor and UE's path tracer.

\noindent\textbf{Assumptions}~~
Throughout the following discussion, we assume the user's Python code is being executed in a particular process on the \emph{Python thread}.
We further assume there is a UE application running in a different process that has been packaged with our SPEAR plugins, and that our plugins are executing the user's UE work as eagerly as possible on the application's \emph{game thread}.

\noindent\textbf{Design Goals}~~
When designing our programming model, we prioritize five main desiderata.
\emph{Expressive.} We want users to be able to express complex graphs of UE work, while also giving them precise control over how their graphs are executed within and across individual UE frames.
\emph{Programmable.} We want to expose as much UE functionality as possible, as directly as possible, without imposing opinionated domain concepts or abstractions on the user.
\emph{Ergonomic.} We want the syntax for calling UE functions and accessing UE variables to be as similar as possible to native Python.
\emph{Extensible.} We want it to be easy for users to expose new C++ functionality.
\emph{Efficient.} We want the speed of programs written in our model to be as close as possible to equivalent programs written directly in C++.

An important non-goal in our model is narrowing the expressive power of UE to make it easier to use for practitioners within a particular subdomain.
There is no doubt that such interfaces are useful, but we believe they are best implemented as optional abstractions on top of our programming model, rather than as required abstractions within our model.

{
\sloppy
\noindent\textbf{Specifying Graphs of Unreal Engine Work}~~
In our programming model, graphs of UE work are specified as transactions (see Figure~\ref{fig:programming_model}).
In particular, the user specifies a transaction by defining a \texttt{\textcolor{myblue}{begin\_frame}} context followed by an \texttt{\textcolor{myblue}{end\_frame}} context.
Within each context, the user specifies a graph of UE work simply by implementing the graph as Python code.
The UE work specified in each \texttt{\textcolor{myblue}{begin\_frame}} context is guaranteed to execute at the beginning of a UE frame, the UE work specified in its corresponding \texttt{\textcolor{myblue}{end\_frame}} context is guaranteed to execute at the end of the \emph{same frame}, and all UE work is guaranteed to execute sequentially on the game thread in the order it is specified in Python.

By default, our programming model guarantees that each transaction executes within a single frame, but we intentionally do not guarantee that consecutive transactions execute on consecutive frames.
In order to implement deterministic stepping in the absence of such a guarantee, the user can begin their UE simulation in a paused state, and in each subsequent transaction unpause the simulation, mutate the game state, and pause the simulation again.
A practical advantage of this approach, as opposed to blocking the game thread between transactions (e.g., CARLA's synchronous mode~\cite{dosovitskiy:2017}), is that the UE application remains responsive to user input at all times.
This interactivity is especially useful when controlling simulations running inside the Unreal Editor, because the editor can be used as a powerful visual debugging tool to navigate around the simulation environment and inspect the game state, even in-between transactions when the simulation is paused.
Alternatively, if desired, the user can explicitly prevent the game thread from advancing in-between transactions by calling \texttt{\textcolor{myblue}{end\_frame(single\_step=True)}}, which instructs the game thread to advance to the beginning of the next frame and wait for the next transaction.
}

\begin{figure}[t]
  \centering
  \includegraphics[width=11.98cm]{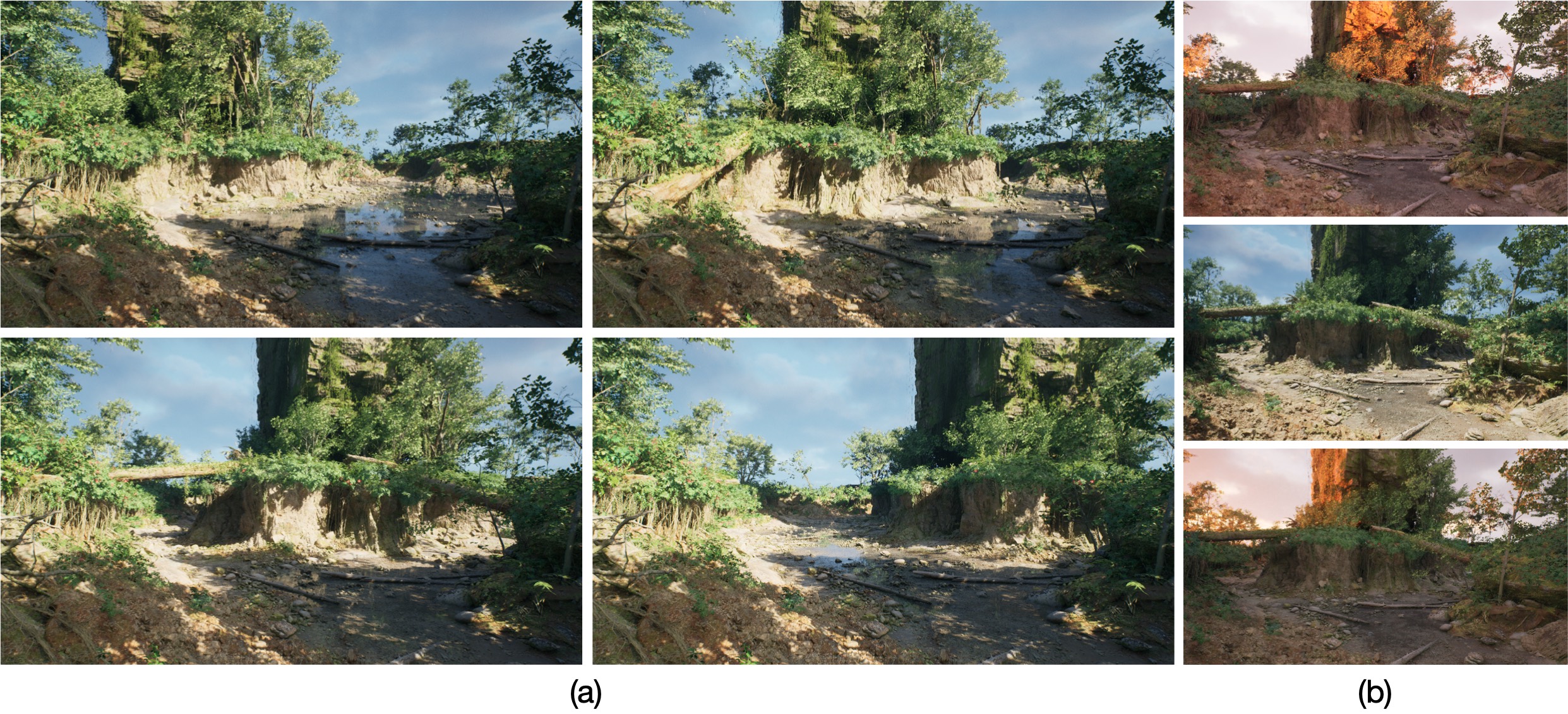}\vspace{-5pt}
  \caption{
We demonstrate the flexibility of SPEAR by using it to programmtically manipulate the \textsc{ElectricDreams} sample project from Epic Games.
\textbf{(a):}
We control UE's procedural content generation (PCG) system by translating the main PCG entity in this scene (the rock structure in the center of each image) from left to right.
Notice how the rock structure automatically harmonizes with the rest of the scene in a convincing way (e.g., the water adjusts around the rock, logs appear and connect with nearby structures), even when it is being driven by our simple programmatic control.
\textbf{(b):} We simulate time-of-day changes by controlling the orientation of the scene's sky light.
}
  \label{fig:electric}
\end{figure}

\noindent\textbf{Calling Functions and Accessing Variables}~~
Given a Python object that represents an underlying UE object (e.g., \texttt{bp\_axes} in Figure \ref{fig:programming_model}), any function or variable on the underlying UE object that is visible to the reflection system can be accessed as though it was a native Python function or attribute on the Python object (e.g., calling the \texttt{\textcolor{myred}{SetActorScale3D}} function and accessing the \texttt{\textcolor{myred}{RootComponent}} variable in Figure~\ref{fig:programming_model}).
Our programming model also supports accessing variables that are nested arbitrarily deeply within structs, e.g., \texttt{\textcolor{myred}{bp\_axes.RootComponent.RelativeLocation.X.get}()} would be a valid expression in Figure~\ref{fig:programming_model}.

\noindent\textbf{Exposing new Functions and Variables}~~
It is trivial to expose new C++ functions and variables to UE's reflection system, and therefore to SPEAR, simply by adding a \texttt{\textcolor{mygreen}{UFUNCTION}} or \texttt{\textcolor{mygreen}{UPROPERTY}} annotation next to the function or variable (e.g., \texttt{\textcolor{mypurple}{MyFunction}} or \texttt{\textcolor{mypurple}{MyProperty}} in Figure~\ref{fig:programming_model}) in any C++ header file, including in header files that are outside of the SPEAR codebase.
An important consequence of this design is that new C++ functionality can be exposed to Python without modifying SPEAR code, which is not possible in any existing UE-based simulator.
No additional registration steps or code wrappers are required to expose new C++ functionality to Python.

\begin{figure}[tb]
  \centering
  \includegraphics[width=12.0cm]{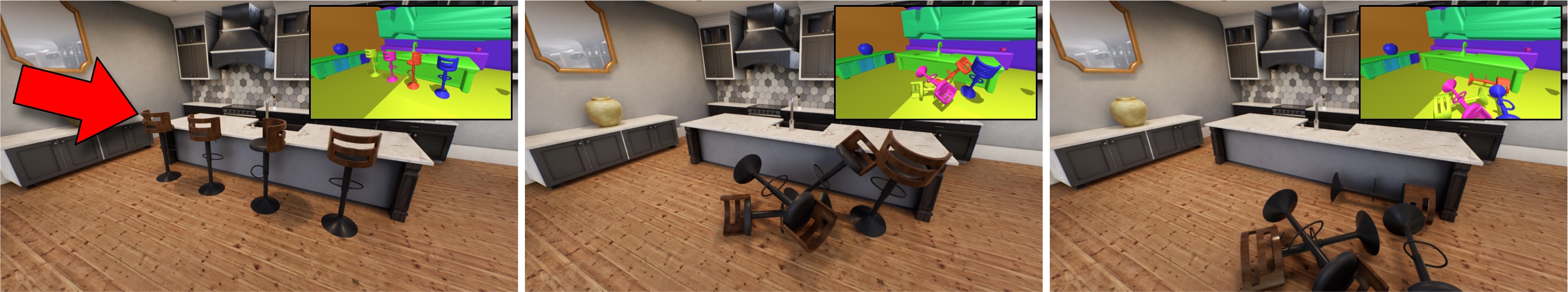}\vspace{-5pt}
  \caption{
SPEAR can be used in co-simulation applications with external physics simulators.
In this application, we interactively control the MuJoCo physics simulator~\cite{todorov:2012} using the default MuJoCo viewer, e.g., by applying a force to the leftmost chair (red arrow).
In real-time as the MuJoCo simulation is running, we query the state of the MuJoCo scene (inset images), and we use SPEAR to update the state of a corresponding UE scene (large images).
}
  \label{fig:mujoco}\vspace{-15pt}
\end{figure}

\noindent\textbf{Synchronizing with the Game Thread}~~
By default, every UE operation in our programming model (e.g., calling a UE function, accessing a UE property) is fully synchronous.
In other words, each operation is guaranteed to finish executing on the game thread before control is returned to the user's Python code.
However, for improved efficiency, we also provide an asynchronous variant for each UE operation in our model (e.g., \texttt{\textcolor{myorange}{async.GetComponentLocation}} in Figure~\ref{fig:programming_model}), which avoids synchronizing with UE and immediately returns a future object.
This future object can be used later in the user's Python program to obtain the return value of the underlying operation (e.g., \texttt{location} in Figure~\ref{fig:programming_model}).
Each asynchronous operation is guaranteed to execute on the game thread in the same order it would have if the user invoked its synchronous counterpart, and getting the return value from an asynchronous operation will not synchronize with the game thread unless the operation is still pending, in which case getting its return value will block until the operation is complete.

When specifying transactions using asynchronous operations, it is possible for the Python thread to execute significantly faster than the game thread.
To prevent these two threads from diverging, we intentionally gate the progress of the Python thread by allowing at most one pending transaction at a time.
Specifically, if work from the user's previous \texttt{\textcolor{myblue}{begin\_frame}} context is still pending, then the next attempt to enter a \texttt{\textcolor{myblue}{begin\_frame}} context will block until the previous work has finished, and likewise for \texttt{\textcolor{myblue}{end\_frame}}.

\noindent\textbf{Executing User Python Code at Native Frame Rates}~~
An important feature of our programming model is that asynchronous operations can be used to completely avoid blocking the game thread, thereby enabling UE to execute user Python code at native frame rates (see Figure~\ref{fig:steady_state}).


\noindent\textbf{Summary}~~
Our programming model is \emph{expressive} because it enables users to specify complex graphs of UE work simply by writing Python code, and because our approach for specifying graphs as transactions gives users precise control over how their graphs are executed.
Our model is \emph{programmable} and \emph{ergonomic} because we expose all reflection-visible UE functionality as native Python functions and attributes.
Our model is \emph{extensible} because it is trivial to expose new C++ functions and variables to the UE reflection system, and therefore to SPEAR.
Finally, our model is \emph{efficient} because it can be used to completely avoid blocking the game thread, thereby enabling UE to execute user Python code at native frame rates.

%% file: 04_system_architecture.tex
\section{System Architecture}
\label{sec:system_architecture}

\begin{figure}[t]
  \centering
  \includegraphics[width=12.0cm]{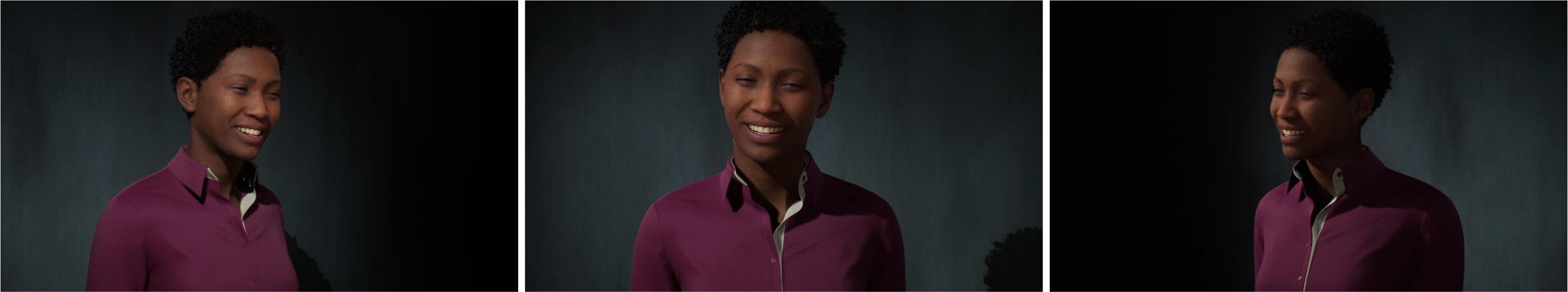}\vspace{-5pt}
  \caption{
We demonstrate the flexibility of our camera sensor by using it to render synchronized multi-view images of a detailed human character in the \nobreak\textsc{MetaHumans} sample project from Epic Games.
}
  \label{fig:metahumans}\vspace{-15pt}
\end{figure}

We implement our programming model using a client-server architecture that is conceptually similar to those found in existing simulators~\cite{bordes:2023,dosovitskiy:2017,gan:2021,juliani:2018,kolve:2017,martinezgonzalez:2019,martinezgonzalez:2021,qiu:2017,shah:2017,zhong:2025}, but with some unique characteristics that are needed for performance and to deliver on the guarantees provided by our programming model.
In our supplementary material, we include additional details about our system architecture, including our high-level Python interface and the implementation of our camera sensor.

\noindent\textbf{Client-Server Interface}~~
We implement a Python client that is responsible for transforming user-defined transactions of UE work into streams of commands that are sent to the server.
Additionally, we implement a C++ server that runs as a plugin within the UE application and responds to commands from the client.
The client communicates with the server over a TCP/IP connection, and it is therefore possible for the client and server to run on different machines.

We implement both sides of our client-server interface in C++ using rpclib~\cite{szelei:2017}, and we implement a Python wrapper for the client side of our interface using nanobind~\cite{jakob:2022}.
This design enables us to implement the server side of our interface as a collection of strongly typed C++ entry points that permit standard containers and our own custom types as arguments and return values.
Additionally, this design enables our client to call server entry points as though they were native C++ functions.
Throughout this discussion, when we say informally that the client sends a command to the server, we mean precisely that the client synchronously calls one of our strongly typed entry points and obtains a strongly typed return value.


\begin{figure}[t]
  \centering
  \includegraphics[width=12.0cm]{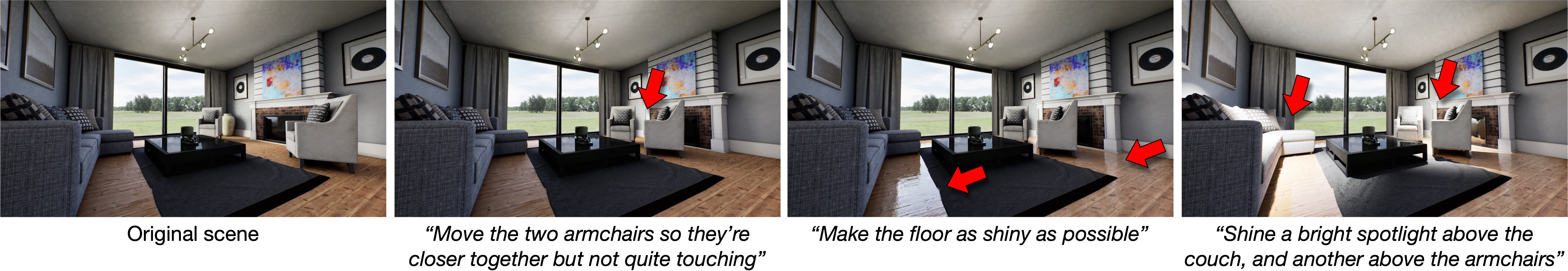}\vspace{-5pt}
  \caption{
We demonstrate the flexibility of SPEAR by using it to implement an agentic natural-language scene editing application, where a vision-and-language coding assistant manipulates a UE scene by iteratively writing SPEAR programs in response to user text prompts.
}
  \label{fig:scene_editing}\vspace{-15pt}
\end{figure}

\noindent\textbf{The Server Thread}~~
In order to support the asynchronous operations in our programming model, we run our server on a separate thread within the UE application, which we refer to as the \emph{server thread}.
We emphasize that the asynchronous operations in our model are asynchronous with respect to the server thread and the game thread, but they are synchronous with respect to the Python thread and the server thread.
This setup simplifies our system architecture, while also ensuring that our server can respond to commands without needing to synchronize with the game thread, in contrast to existing UE plugins (e.g.,~\cite{qiu:2017,zhong:2025}).

Because our server runs on its own thread, it needs a well-defined way of interacting with the game thread, since nearly all UE work specified by the client must access the game thread at some point.
With this goal in mind, we implement a thread-safe task queuing system.
In particular, we maintain two separate thread-safe queues for \texttt{\textcolor{mygrey}{begin\_frame}} and \texttt{\textcolor{mygrey}{end\_frame}} work respectively.
Whenever an entry point is called that needs to access the game thread, the server places a task in one of these queues, and the queues are drained by the game thread at the beginning and end of each UE frame in a way that respects the execution ordering guarantees in our programming model.

\noindent\textbf{Synchronizing Progress Across Threads}~~
In our system architecture, we rely on the server to synchronize the progress of both the Python thread and the game thread according to the rules of our programming model.
To perform this synchronization, the server must track the state of the user's Python code as it enters and exits \texttt{\textcolor{mygrey}{begin\_frame}} and \texttt{\textcolor{mygrey}{end\_frame}} contexts.

We keep the server up-to-date by implementing the following logic in our \texttt{\textcolor{mygrey}{begin\_frame}} and \texttt{\textcolor{mygrey}{end\_frame}} context managers~\cite{vanrossum:2005} on the client.
When the user's Python code enters a \texttt{\textcolor{mygrey}{begin\_frame}} context, the client sends a \texttt{\textcolor{mygreen}{begin}} command to the server; when the user's Python code leaves a \texttt{\textcolor{mygrey}{begin\_frame}} context, the client sends an \texttt{\textcolor{mybrightblue}{execute}} command to the server; and when the user's Python code leaves an \texttt{\textcolor{mygrey}{end\_frame}} context, the client sends an \texttt{\textcolor{mypurple}{end}} command to the server.
The \texttt{\textcolor{mygreen}{begin}} and \texttt{\textcolor{mybrightblue}{execute}} commands will block if the server determines that the user's Python code is too far ahead of the game thread (see Figure~\ref{fig:steady_state}).

\noindent\textbf{Hand-Crafted Server Entry Points}~~
We implement 193 hand-crafted server entry points to expose various UE functions that are not visible to the reflection system, as well as the reflection system itself.
We implement each synchronous variant manually, and we generate its asynchronous variant automatically using template metaprogramming~\cite{vandevoorde:2017}.
This technique leverages the fact that each synchronous variant is implemented as a C++ function, rather than as a text-based command that must be parsed and dispatched imperatively (e.g., as in~\cite{qiu:2017,zhong:2025}).
When the user's Python code is inside a \texttt{\textcolor{mygrey}{begin\_frame}} or \texttt{\textcolor{mygrey}{end\_frame}} context, all of our server entry points can be called as Python functions (e.g., \texttt{\textcolor{myred}{load\_class}} and \texttt{\textcolor{myred}{spawn\_actor}} in Figure~\ref{fig:programming_model}).
Internally, our Python client routes each such function call through our nanobind wrapper and ultimately to the appropriate entry point on the server.


\noindent\textbf{Exposing the Reflection System}~~
Roughly 75\% of our hand-crafted server entry points are dedicated to exposing some aspect of the UE reflection system.
Together, these entry points enable our Python client to find classes, call functions, and manipulate variables on objects dynamically at runtime using strings as keys, i.e., without requiring hand-crafted wrapper code for each class, function, and variable.
Through this approach, we ultimately expose over 14K UE functions and 53K UE properties to Python.

When exposing UE's reflection system to Python, we leverage the fact that UE imposes some restrictions on the types that it considers to be \emph{reflectable}~\cite{unrealengine:2026:reflection}.
In other words, only certain types are allowed in reflected function signatures and member variables (e.g., primitive types, strings, pointers to UE objects, some containers, \texttt{enums}, and \texttt{structs} that are recursively composed of types from this set).
These restrictions enable UE to automatically serialize and deserialize to and from JSON any variable whose type is reflectable, which is especially useful for our purposes, because Python dictionaries can also be automatically serialized and deserialized to and from JSON.
The structural alignment between reflectable types in UE and dictionaries in Python enables us to use dictionaries as a universal representation for getting and setting variables, as well as passing arguments and return values to and from functions in user Python code (e.g., \texttt{NewScale3D} in Figure~\ref{fig:programming_model}).

\noindent\textbf{NumPy Interoperability}~~
When passing large blocks of data (e.g., high-resolution images) to and from UE functions, our approach for representing arguments and return values as JSON strings can be inefficient.
We address this limitation by implementing a dispatching system for invoking custom functions on UE objects, which supports NumPy arrays as arguments and return values and marshals them efficiently.
In our system, any UE object that has a hierarchy of child components (see~\cite{unrealengine:2026:components} for a more detailed discussion of UE's component system) can define custom functions by inserting a special child component into its hierarchy and binding named functions to this component at runtime.
We refer to these named functions as \textsc{SpFunctions}, and they can be called from user Python code as though they were reflection-visible UE functions, i.e., they appear to user Python code as native Python functions in our programming model.
Defining an \textsc{SpFunction} in C++ code is trivial, requiring a similar amount of effort as exposing a function to UE's reflection system (see our supplementary material for a code example).

For simplicity, we restrict the signature of \textsc{SpFunctions} in the following way.
All \textsc{SpFunctions} must take as input, and return as output, three distinct objects: (1) a collection of named data arrays with shape and type metadata; (2) a collection of named UE objects represented as JSON strings; and (3) a string that can be used to encode miscellaneous user data.
When a call to an \textsc{SpFunction} reaches our client-server boundary, any NumPy arrays that are given as arguments are mapped to our named data array representation, and any data arrays that are returned as output are mapped back to NumPy arrays.

\setlength{\tabcolsep}{4pt}
\begin{table}[t]
\centering
\begin{tabular}{@{}llllll@{}}
\toprule
Simulator    & Functions    & Variables    & UE functions    & UE variables    & Lines of code   \\
\midrule
AirSim       & 92           & 189          & 0               & 0               & 144,536         \\
CARLA        & \textbf{465} & \textbf{508} & 0               & 0               & 150,502         \\
UnrealCV+    & 56           & 0            & 747             & 8,721           & \textbf{11,301} \\
SPEAR (ours) & 193          & 67           & \textbf{14,485} & \textbf{53,537} & 27,193          \\
\bottomrule
\end{tabular}
\vspace{5pt}
\caption{
\textbf{Comparison of programmable functionality across UE simulators.}
We report the number of hand-crafted functions and variables each simulator~\cite{dosovitskiy:2017,qiu:2017,shah:2017,zhong:2025} provides, as well as the number of underlying UE functions and variables it exposes (higher is better).
For reference, we also report the total lines of Python and C++ code in each simulator codebase (lower is better).
SPEAR provides an order of magnitude more programmable functionality than existing UE-based simulators while maintaining a modest code footprint.
}
\label{tbl:api_comparison}\vspace{-15pt}
\end{table}

\noindent\textbf{Interprocess Shared Memory}~~
We further optimize the efficiency of our system using interprocess shared memory~\cite{schaeling:2011}, or simply \emph{shared memory}.
Using shared memory, we can, e.g., transfer rendered images from the GPU directly into a user's NumPy array with no additional data copying.

We integrate shared memory into our system by implementing a C++ interface for allocating and deallocating shared memory regions, and we expose this interface to Python through a set of hand-crafted server entry points.
Owing to this interface, user Python code can optionally allocate a shared memory region, create a NumPy array that is backed by that region, and pass the NumPy array to an \textsc{SpFunction}.
Importantly, when our server calls an \textsc{SpFunction} implementation, we internally resolve references to shared memory for all arguments prior to calling the implementation.
As a result, the implementation remains completely decoupled from the caller's decision to use shared memory for arguments.
We implement a similar strategy when returning data from an \textsc{SpFunction}, so the user's Python code remains decoupled from the implementation's decision to use shared memory for return values.

%% file: 05_results.tex
\vspace{-5pt}
\section{Results}
\label{sec:results}

\setlength{\tabcolsep}{4pt}
\begin{table}[t]
\centering
\begin{minipage}{0.57\linewidth}
\centering
\begin{tabular}{@{}lll@{}}
\toprule
Configuration                                        & Time (ms)     & FPS           \\
\cmidrule{1-3}
Standalone                                           & 7.7           & 129.9         \\
Standalone $+$ extra work                            & 17.7          & 56.5          \\
\cmidrule{1-3}
UnrealCV+                                            & 286.9         & 3.5           \\
\cmidrule{1-3}
SPEAR (ours)                                         &               &               \\
~~~~$\times$     async ~ $\times$     shared mem~~   & 40.5          & 24.7          \\
~~~~$\times$     async ~ $\checkmark$ shared mem~~   & 31.6          & 31.7          \\
~~~~$\checkmark$ async ~ $\times$     shared mem~~   & 37.3          & 26.8          \\
~~~~$\checkmark$ async ~ $\checkmark$ shared mem~~   & \textbf{17.8} & \textbf{56.2} \\
\cmidrule{1-3}
SPEAR (ours)                                         &               &               \\
~~~~1 frame rendering latency~~                      & 15.4          & 64.8          \\
~~~~2 frame rendering latency~~                      & \textbf{13.6} & \textbf{73.4} \\
\bottomrule
\end{tabular}
\end{minipage}
~~
\begin{minipage}{0.33\linewidth}
\centering
\includegraphics[width=1.65in]{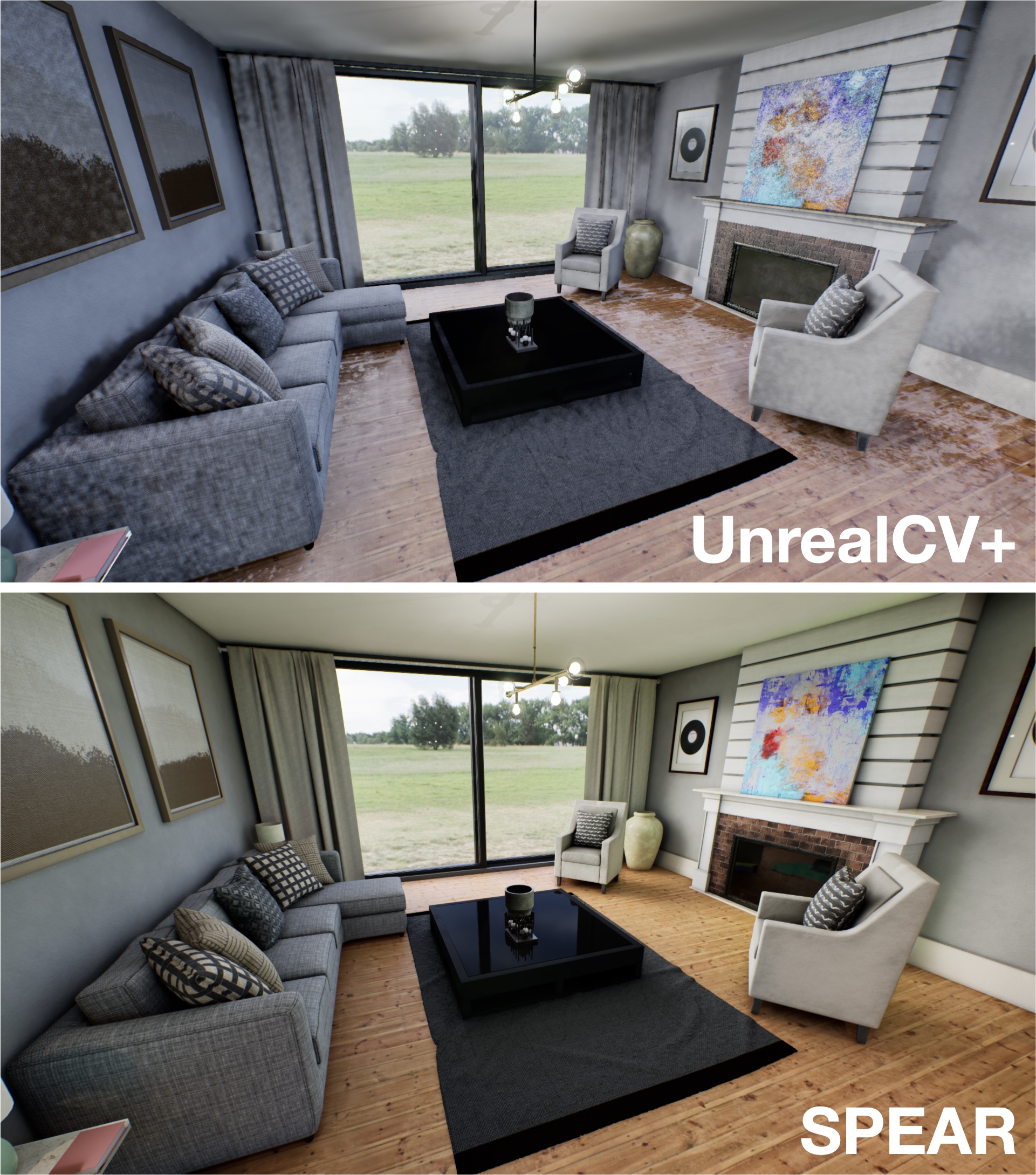}
\end{minipage}
\vspace{5pt}
\caption{
\textbf{Comparison of rendering performance under various configurations} when rendering the image on the right at 1920$\times$1080 resolution.
For each configuration, we measure the total end-to-end time required to deliver a rendered image to a user's Python program, and we report frame time (ms) and frames per second (FPS).
As baselines, we include a standalone UE executable that does not communicate with Python, a standalone executable that does not communicate with Python but does the extra UE work necessary to do so (e.g., rendering an extra view of the scene to an off-screen buffer), and UnrealCV+~\cite{qiu:2017,zhong:2025}.
We report the performance of SPEAR with and without asynchronous communication and shared memory, and with different amounts of rendering latency.
We find that asynchronous communication and shared memory both improve performance, as does increasing rendering latency, and that SPEAR is 9--21$\times$ faster than UnrealCV+ while also supporting more photorealistic rendering.
}
\label{tbl:ablation}\vspace{-20pt}
\end{table}

\setlength{\tabcolsep}{4pt}
\begin{table}[t]
\centering
\begin{minipage}{0.43\linewidth}
\centering
\begin{tabular}{@{}lll@{}}
\toprule
Simulator        & Time (ms)     & FPS               \\
\cmidrule{1-3}
\multicolumn{3}{@{}l@{}}{0 frames rendering latency} \\
~~~~AirSim       & 379.4         & 2.6               \\
~~~~SPEAR (ours) & \textbf{31.0} & \textbf{32.3}     \\
\cmidrule{1-3}
\multicolumn{3}{@{}l@{}}{2 frames rendering latency} \\
~~~~CARLA        & 30.6          & 32.7              \\
~~~~SPEAR (ours) & \textbf{27.0} & \textbf{37.1}     \\
\bottomrule
\end{tabular}
\end{minipage}
~
\begin{minipage}{0.48\linewidth}
\centering
\includegraphics[width=2.3in]{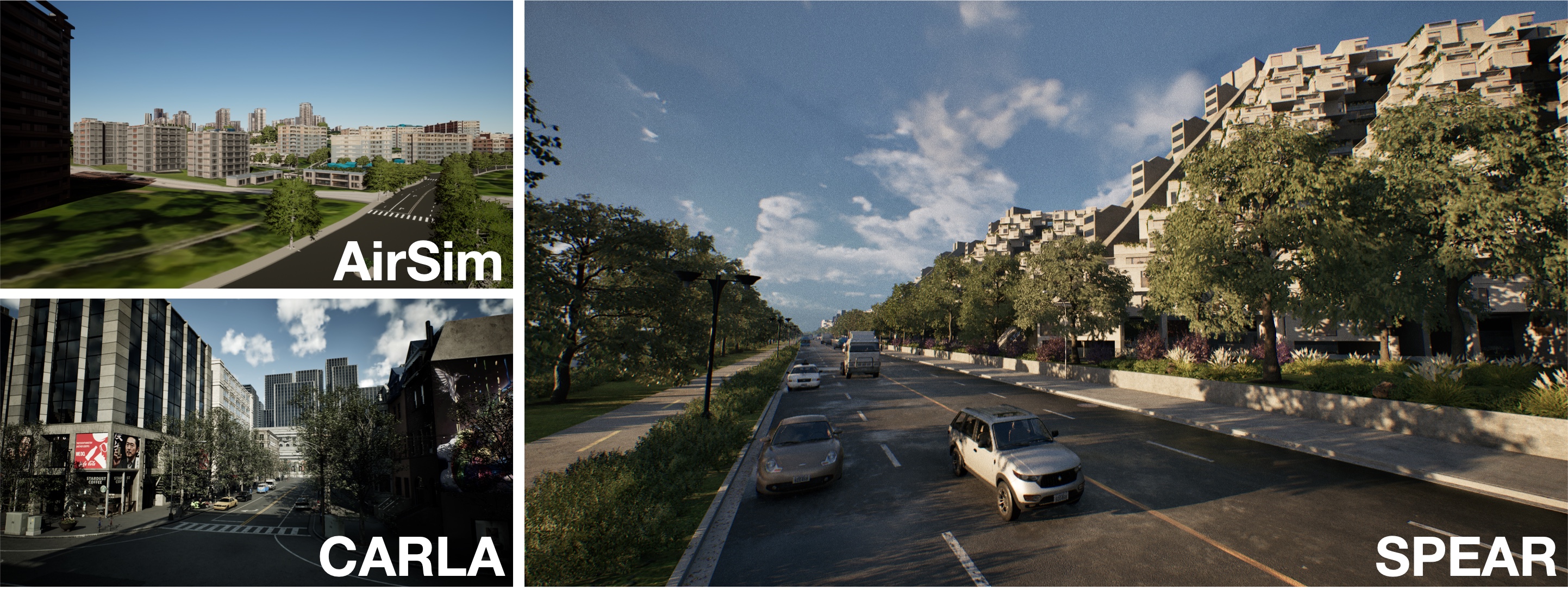}
\end{minipage}
\vspace{5pt}
\caption{
\textbf{Comparison of rendering performance across simulators} when rendering the images on the right at 1920$\times$1080 resolution.
For each simulator~\cite{dosovitskiy:2017,shah:2017}, we measure the total end-to-end time required to deliver a rendered image to a user's Python program, and we report frame time (ms) and frames per second (FPS).
We ensure the rendering speed of each simulator is as similar as possible when running as a standalone executable that does not communicate with Python (SPEAR:~89~FPS;~CARLA:~90~FPS;~AirSim:~93~FPS).
Under these conditions, any differences we observe when measuring total end-to-end rendering time are attributable to differences in communication overhead.
We find that SPEAR is 12$\times$ faster than AirSim, and is 10\% faster than CARLA under matched amounts of rendering latency.
}
\label{tbl:timing}\vspace{-30pt}
\end{table}

\noindent\textbf{Example Applications}~~
We demonstrate the utility of SPEAR through a diverse collection of example applications (see Figures~\ref{fig:control_sample_projects},~\ref{fig:electric},~\ref{fig:mujoco},~\ref{fig:metahumans}, and~\ref{fig:scene_editing}).
We include several additional applications in our supplementary material, and all of our applications are included in our public code repository.

Every example application we show in our figures and videos would not be possible to implement in any existing UE-based simulator because existing simulators do not expose the necessary functionality (e.g., to interact with UE's procedural content system, or control UE's path tracer, etc).
Additionally, nearly all of the functionality required in these applications is already exposed through UE's reflection system, and is therefore exposed automatically in SPEAR.
In summary, SPEAR exposes more functions than existing simulators (i.e., SPEAR is more programmable), and this makes it possible to implement a wider range of programs than would be possible otherwise.
See our supplementary material for a more detailed discussion.


\noindent\textbf{Programmable functionality}~~
In Table~\ref{tbl:api_comparison}, we compare programmable functionality and code size across UE-based simulators.
For each baseline simulator, we include all publicly documented Python functions in our count of its hand-crafted functions.
If a simulator can be configured via some kind of global settings file, we include all documented settings in our count of its hand-crafted variables.
UnrealCV+~\cite{qiu:2017,zhong:2025} uses a text-based command interface, so we include all publicly documented text commands in our count of its hand-crafted functions.
Both UnrealCV+ and SPEAR enable users to execute UE console commands and access UE console variables~\cite{unrealengine:2026:console}, so we include these functions and variables in our counts for both simulators.
For SPEAR, we report the number of unique server entry points and YAML configuration options as our count of hand-crafted functions and variables.
To count the number of UE functions and variables we expose in SPEAR, we use our reflection interface to iterate over all reflection-visible UE functions and variables (i.e., all \texttt{\textcolor{mygreen}{UFUNCTIONS}} and \texttt{\textcolor{mygreen}{UPROPERTIES}}).

\noindent\textbf{Rendering Performance}~~
We perform all performance comparisons on a Windows 11 desktop workstation with an NVIDIA GeForce RTX 4090 GPU, a 4.5 GHz AMD Ryzen 9 processor with 16 physical cores, and 192 GB of memory.
We include additional performance comparisons in our supplementary material.

In Table~\ref{tbl:ablation}, we compare rendering performance relative to UnrealCV+~\cite{qiu:2017,zhong:2025} and a standalone executable that does not communicate with Python.
For this experiment, we build one executable that includes our SPEAR plugins, and another that includes the UnrealCV+ plugin, in an otherwise identical UE project.
This approach enables us to evaluate the rendering performance of both plugins under precisely controlled conditions (e.g., same view of the same scene with all of the same project settings).
The SPEAR camera sensor is designed to incur a user-configurable amount of rendering latency in exchange for increased throughput (see our supplementary material for details).
All rows in Table~\ref{tbl:ablation} are configured to incur 0 frames of latency unless otherwise noted, and all rows that incur 1 or 2 frames of rendering latency use shared memory.

In Table~\ref{tbl:timing}, we compare rendering performance across UE-based simulators.
For this experiment, it is not straightforward to render the same scene in all simulators, because AirSim does not distribute any non-trivial environments in source form, and CARLA environments are tightly coupled to the CARLA source code.
So we render a different scene in each simulator, but we ensure the rendering speed of each simulator is as similar as possible when running as a standalone executable that does not communicate with Python (SPEAR:~89~FPS;~CARLA:~90~FPS;~AirSim:~93~FPS).
Under these conditions, any differences we observe when measuring total end-to-end rendering time with Python are attributable to differences in communication overhead.

For AirSim and CARLA, we use their default environments, and we select views of each environment that render at roughly the same speed.
For SPEAR, we use the \textsc{HillsideSample} project from Epic Games, because it depicts a city-scale urban outdoor environment that is qualitatively similar to the default environments in AirSim and CARLA.
However, \textsc{HillsideSample} is designed to be a challenging next-generation showcase for UE's latest rendering features, and therefore it renders noticeably slower than the default AirSim and CARLA environments.
To account for this discrepancy, we remove distant background objects and adjust rendering quality settings until we achieve roughly the same standalone rendering speed across all simulators, while giving a slight advantage to AirSim and CARLA (see above).
AirSim and CARLA incur 0 and 2 frames of rendering latency respectively, so we report the performance of SPEAR with each of these latencies to ensure fair comparisons, and we enable shared memory.

%% file: 06_conclusions.tex
\vspace{-5pt}
\section{Conclusions}
\label{sec:conclusions}

In this work, we introduced SPEAR: A Simulator for Photorealistic Embodied AI Research.
At its core, SPEAR is a Python library that can programmatically control any Unreal Engine (UE) application using an expressive high-level programming model.
The SPEAR programming model exposes over 14K unique UE functions to Python (an order of magnitude more programmable functionality than existing UE-based simulators), and enables photorealistic rendering at over 150 megapixels per second (an order of magnitude faster than existing UE plugins).
Additionally, the default SPEAR camera sensor is highly configurable, and provides ground truth image modalities that are not available in any existing UE-based simulator.
We demonstrated the utility of SPEAR through a diverse collection of example applications, including controlling multiple embodied agents in a photorealistic city-scale environment, controlling UE's procedural content generation systems, rendering multi-view images of detailed human faces, coordinating an interactive co-simulation with MuJoCo, and editing scenes with natural language via an AI coding assistant.

We believe that SPEAR could become a foundational data engine in computer vision, robotics, and embodied AI.
In the near future, SPEAR could be used to train agile robots in city-scale photorealistic environments, as well as interactive world models that understand the dynamics and spatial structure of the physical world.
By providing programmability via small Python programs, and observability via fast photorealistic rendering, SPEAR is ideally positioned as a bridge to connect internet-scale vision-and-language models to state-of-the-art virtual worlds in the Unreal Engine.
In turn, this could lead to new forms of AI-assisted content creation and personalized entertainment, as well as new virtual laboratories for studying the foundations of spatial intelligence.

%% file: 07_acknowledgments.tex
\vspace{-5pt}
\section*{Acknowledgments}
\label{sec:acknowledgments}

We thank Ahlam Laatiki for creating the default SPEAR apartment scene; Epic Games for making the sample projects that we use throughout this paper available for download; Songyou Peng for his excellent guidance on crafting our rebuttal; and Kevin Blackburn-Matzen, Samarth Brahmbhatt, Marti Ferragut Galtes, and David Hafner for their assistance with early prototyping.

%% file: main.bib
@String(CVPR  = {IEEE Conf. Comput. Vis. Pattern Recog.})

@String(ICCV  = {Int. Conf. Comput. Vis.})

@String(ECCV  = {Eur. Conf. Comput. Vis.})

@String(NeurIPS = {Adv. Neural Inform. Process. Syst.})

@String(ICML  = {Int. Conf. Mach. Learn.})

@String(ICLR  = {Int. Conf. Learn. Represent.})

@String(IJCAI = {IJCAI})

@String(CVPR  = {CVPR})

@String(ICCV  = {ICCV})

@String(ECCV  = {ECCV})

@String(NeurIPS = {NeurIPS})

@String(ICML  = {ICML})

@String(ICLR  = {ICLR})

@book{vandevoorde:2017,
    author       = {David Vandevoorde AND Nicolai M. Josuttis AND Douglas Gregor},
    title        = {{C++} Templates: {T}he Complete Guide},
    publisher    = {Addison-Wesley Professional},
    year         = {2017},
    edition      = {2}
}

@book{schaeling:2011,
  author    = {Boris Sch{\"a}ling},
  title     = {The {Boost} {C++} Libraries},
  publisher = {XML Press},
  year      = {2011}
}

@article{bellemare:2013,
    author       = {Marc G. Bellemare AND Yavar Naddaf AND Joel Veness AND Michael Bowling},
    title        = {The Arcade Learning Environment: {A}n Evaluation Platform for General Agents},
    journal      = {Journal of Artificial Intelligence Research},
    volume       = {47},
    number       = {1},
    year         = {2013}
}

@article{mnih:2015,
    author       = {Volodymyr Mnih AND Koray Kavukcuoglu AND David Silver AND Andrei A. Rusu AND Joel Veness AND Marc G. Bellemare AND Alex Graves AND Martin Riedmiller AND Andreas K. Fidjeland AND Georg Ostrovski AND Stig Petersen AND Charles Beattie AND Amir Sadik AND Ioannis Antonoglou AND Helen King AND Dharshan Kumaran AND Daan Wierstra AND Shane Legg AND Demis Hassabis},
    title        = {Human-Level Control through Deep Reinforcement Learning},
    journal      = {Nature},
    volume       = {518},
    number       = {7540},
    year         = {2015}
}

@article{zhou:2018,
    author       = {Brady Zhou AND Philipp Kr{\"a}henb{\"u}hl AND Vladlen Koltun},
    title        = {Does Computer Vision Matter for Action?},
    journal      = {Science Robotics},
    volume       = {4},
    number       = {30},
    year         = {2018}
}

@article{hwangbo:2019,
    author       = {Jemin Hwangbo AND Joonho Lee AND Alexey Dosovitskiy AND Dario Bellicoso AND Vassilios Tsounis AND Vladlen Koltun AND Marco Hutter},
    title        = {Learning Agile and Dynamic Motor Skills for Legged Robots},
    journal      = {Science Robotics},
    volume       = {4},
    number       = {26},
    year         = {2019}
}

@article{jaderberg:2019,
    author       = {Max Jaderberg AND Wojciech M. Czarnecki AND Iain Dunning AND Luke Marris AND Guy Lever AND Antonio Garcia Casta{\~n}eda AND Charles Beattie AND Neil C. Rabinowitz AND Ari S. Morcos AND Avraham Ruderman AND Nicolas Sonnerat AND Tim Green AND Louise Deason AND Joel Z. Leibo AND David Silver AND Demis Hassabis AND Koray Kavukcuoglu AND Thore Graepel},
    title        = {Human-level Performance in {3D} Multiplayer Games with Population-Based Deep Reinforcement Learning},
    journal      = {Science},
    volume       = {364},
    number       = {6443},
    year         = {2019}
}

@article{vinyals:2019,
    author       = {Oriol Vinyals AND Igor Babuschkin AND Wojciech M. Czarnecki AND Micha{\"e}l Mathieu AND Andrew Dudzik AND Junyoung Chung AND David H. Choi AND Richard Powell AND Timo Ewalds AND Petko Georgiev AND Junhyuk Oh AND Dan Horgan AND Manuel Kroiss AND Ivo Danihelka AND Aja Huang AND Laurent Sifre AND Trevor Cai AND John P. Agapiou AND Max Jaderberg AND Alexander S. Vezhnevets AND R{\'e}mi Leblond AND Tobias Pohlen AND Valentin Dalibard AND David Budden AND Yury Sulsky AND James Molloy AND Tom L. Paine AND Caglar Gulcehre AND Ziyu Wang AND Tobias Pfaff AND Yuhuai Wu AND Roman Ring AND Dani Yogatama AND Dario W{\"u}nsch AND Katrina McKinney AND Oliver Smith AND Tom Schaul AND Timothy Lillicrap AND Koray Kavukcuoglu AND Demis Hassabis AND Chris Apps AND David Silver},
    title        = {Grandmaster Level in {StarCraft} {II} using Multi-Agent Reinforcement Learning},
    journal      = {Nature},
    volume       = {575},
    year         = {2019}
}

@article{lee:2020,
    author       = {Joonho Lee AND Jemin Hwangbo AND Lorenz Wellhausen AND Vladlen Koltun AND Marco Hutter},
    title        = {Learning Quadrupedal Locomotion over Challenging Terrain},
    journal      = {Science Robotics},
    volume       = {5},
    number       = {47},
    year         = {2020}
}

@article{loquercio:2020,
    author       = {Antonio Loquercio AND Elia Kaufmann AND Ren{\'e} Ranftl AND Alexey Dosovitskiy AND Vladlen Koltun AND Davide Scaramuzza},
    title        = {Deep Drone Racing: {F}rom Simulation to Reality with Domain Randomization},
    journal      = {Transactions on Robotics},
    volume       = {36},
    number       = {1},
    year         = {2020}
}

@article{loquercio:2021,
    author       = {Antonio Loquercio AND Elia Kaufmann AND Ren{\'e} Ranftl AND Matthias M{\"u}ller AND Vladlen Koltun AND Davide Scaramuzza},
    title        = {Learning High-Speed Flight in the Wild},
    journal      = {Science Robotics},
    volume       = {6},
    number       = {59},
    year         = {2021}
}

@article{martinezgonzalez:2019,
    author       = {Pablo Martinez-Gonzalez AND Sergiu Oprea AND Alberto Garcia-Garcia AND Alvaro Jover-Alvarez AND Sergio Orts-Escolano AND Jose Garcia-Rodriguez},
    title        = {{UnrealROX}: {A}n eXtremely Photorealistic Virtual Reality Environment for Robotics Simulations and Synthetic Data Generation},
    journal      = {Virtual Reality},
    volume       = {24},
    year         = {2019}
}

@article{miki:2022,
    author       = {Takahiro Miki AND Joonho Lee AND Jemin Hwangbo AND Lorenz Wellhausen AND Vladlen Koltun AND Marco Hutter},
    title        = {Learning Robust Perceptive Locomotion for Quadrupedal Robots in the Wild},
    journal      = {Science Robotics},
    volume       = {7},
    number       = {62},
    year         = {2022}
}

@article{wurman:2022,
    author       = {Peter R. Wurman AND Samuel Barrett AND Kenta Kawamoto AND James MacGlashan AND Kaushik Subramanian AND Thomas J. Walsh AND Roberto Capobianco AND Alisa Devlic AND Franziska Eckert AND Florian Fuchs AND Leilani Gilpin AND Piyush Khandelwal AND Varun Kompella AND HaoChih Lin AND Patrick MacAlpine AND Declan Oller AND Takuma Seno AND Craig Sherstan AND Michael D. Thomure AND Houmehr Aghabozorgi AND Leon Barrett AND Rory Douglas AND Dion Whitehead AND Peter D{\"u}rr AND Peter Stone AND Michael Spranger AND Hiroaki Kitano},
    title        = {Outracing Champion {Gran Turismo} Drivers with Deep Reinforcement Learning},
    journal      = {Nature},
    volume       = {602},
    year         = {2022}
}

@article{kaufmann:2023,
    author       = {Elia Kaufmann AND Leonard Bauersfeld AND Antonio Loquercio AND Matthias M{\"u}ller AND Vladlen Koltun AND Davide Scaramuzza},
    title        = {Champion-level Drone Racing using Deep Reinforcement Learning},
    journal      = {Nature},
    volume       = {620},
    number       = {7976},
    year         = {2023}
}

@inproceedings{maes:1987,
    author       = {Pattie Maes},
    title        = {Concepts and Experiments in Computational Reflection},
    booktitle    = {OOPSLA 1987}
}

@inproceedings{todorov:2012,
    author       = {Emanuel Todorov AND Tom Erez AND Yuval Tassa},
    title        = {{MuJoCo}: {A} Physics Engine for Model-Based Control},
    booktitle    = {IROS 2012}
}

@inproceedings{johnson:2016,
    author       = {Matthew Johnson AND Katja Hofmann AND Tim Hutton AND David Bignell},
    title        = {The {Malmo} Platform for Artificial Intelligence Experimentation},
    booktitle    = {IJCAI 2016}
}

@inproceedings{kempka:2016,
    author       = {Michał Kempka AND Marek Wydmuch AND Grzegorz Runc AND Jakub Toczek AND Wojciech Jaśkowski},
    title        = {{ViZDoom}: {A} {Doom}-Based {AI} Research Platform for Visual Reinforcement Learning},
    booktitle    = {Computational Intelligence and Games 2016}
}

@inproceedings{lerer:2016,
    author       = {Adam Lerer AND Sam Gross AND Rob Fergus},
    title        = {Learning Physical Intuition of Block Towers by Example},
    booktitle    = {ICML 2016}
}

@inproceedings{richter:2016,
    author       = {Stephan Richter AND Vibhav Vineet AND Stefan Roth AND Vladlen Koltun},
    title        = {Playing for Data: {G}round Truth from Computer Games},
    booktitle    = {ECCV 2016}
}

@inproceedings{dosovitskiy:2017,
    author       = {Alexey Dosovitskiy AND German Ros AND Felipe Codevilla AND Antonio Lopez AND Vladlen Koltun},
    title        = {{CARLA}: {A}n Open Urban Driving Simulator},
    booktitle    = {CoRL 2017}
}

@inproceedings{qiu:2017,
    author       = {Weichao Qiu AND Fangwei Zhong AND Yi Zhang AND Siyuan Qiao AND Zihao Xiao AND Tae Soo Kim AND Yizhou Wang},
    title        = {{UnrealCV}: {V}irtual Worlds for Computer Vision},
    booktitle    = {Multimedia 2017}
}

@inproceedings{richter:2017,
    author       = {Stephan R. Richter AND Zeeshan Hayder AND Vladlen Koltun},
    title        = {Playing for Benchmarks},
    booktitle    = {ICCV 2017}
}

@inproceedings{shah:2017,
    author       = {Shital Shah AND Debadeepta Dey AND Chris Lovett AND Ashish Kapoor},
    title        = {{AirSim}: {H}igh-Fidelity Visual and Physical Simulation for Autonomous Vehicles},
    booktitle    = {Field and Service Robotics 2017}
}

@inproceedings{kaufmann:2018,
    author       = {Elia Kaufmann AND Antonio Loquercio AND Rene Ranftl AND Alexey Dosovitskiy AND Vladlen Koltun AND Davide Scaramuzza},
    title        = {Deep Drone Racing: {L}earning Agile Flight in Dynamic Environments},
    booktitle    = {CoRL 2018}
}

@inproceedings{krahenbuhl:2018,
    author       = {Philipp Kr\"ahenb\"uhl},
    title        = {Free Supervision from Video Games},
    booktitle    = {CVPR 2018}
}

@inproceedings{muller:2018,
    author       = {Matthias M{\"u}ller AND Alexey Dosovitskiy AND Bernard Ghanem AND Vladlen Koltun},
    title        = {Driving Policy Transfer via Modularity and Abstraction},
    booktitle    = {CoRL 2018}
}

@inproceedings{xia:2018,
    author       = {Fei Xia AND Amir R. Zamir AND Zhi-Yang He AND Alexander Sax AND Jitendra Malik AND Silvio Savarese},
    title        = {{Gibson Env}: {R}eal-World Perception for Embodied Agents},
    booktitle    = {CVPR 2018}
}

@inproceedings{savva:2019,
    author       = {Manolis Savva AND Abhishek Kadian AND Oleksandr Maksymets AND Yili Zhao AND Erik Wijmans AND Bhavana Jain AND Julian Straub AND Jia Liu AND Vladlen Koltun AND Jitendra Malik AND Devi Parikh AND Dhruv Batra},
    title        = {Habitat: {A} Platform for Embodied {AI} Research},
    booktitle    = {ICCV 2019}
}

@inproceedings{kaufmann:2020,
    author       = {Elia Kaufmann AND Antonio Loquercio AND Rene Ranftl AND Matthias M{\"u}ller AND Vladlen Koltun AND Davide Scaramuzza},
    title        = {Deep Drone Acrobatics},
    booktitle    = {RSS 2020}
}

@inproceedings{xiang:2020,
    author       = {Fanbo Xiang AND Yuzhe Qin AND Kaichun Mo AND Yikuan Xia AND Hao Zhu AND Fangchen Liu AND Minghua Liu AND Hanxiao Jiang AND Yifu Yuan AND He Wang AND Li Yi AND Angel X. Chang AND Leonidas J. Guibas AND Hao Su},
    title        = {{SAPIEN}: {A} SimulAted Part-based Interactive ENvironment},
    booktitle    = {CVPR 2020}
}

@inproceedings{gan:2021,
    author       = {Chuang Gan AND Jeremy Schwartz AND Seth Alter AND Damian Mrowca AND Martin Schrimpf AND James Traer AND Julian De Freitas AND Jonas Kubilius AND Abhishek Bhandwaldar AND Nick Haber AND Megumi Sano AND Kuno Kim AND Elias Wang AND Michael Lingelbach AND Aidan Curtis AND Kevin Feigelis AND Daniel M. Bear AND Dan Gutfreund AND David Cox AND Antonio Torralba AND James J. DiCarlo AND Joshua B. Tenenbaum AND Josh H. McDermott AND Daniel L. K. Yamins},
    title        = {{ThreeDWorld}: {A} Platform for Interactive Multi-Modal Physical Simulation},
    booktitle    = {NeurIPS 2021 Datasets and Benchmarks Track}
}

@inproceedings{karis:2021,
    author       = {Brian Karis AND Rune Stubbe AND Graham Wihlidal},
    title        = {A Deep Dive into {N}anite Virtualized Geometry},
    booktitle    = {SIGGRAPH 2021 Course on Advances in Real-Time Rendering and Games}
}

@inproceedings{li:2021,
    author       = {Chengshu Li AND Fei Xia AND Roberto Mart{\'i}n-Mart{\'i}n AND Michael Lingelbach AND Sanjana Srivastava AND Bokui Shen AND Kent Elliott Vainio AND Cem Gokmen AND Gokul Dharan AND Tanish Jain AND Andrey Kurenkov AND C. Karen Liu AND Hyowon Gweon AND Jiajun Wu AND Li Fei-Fei AND Silvio Savarese},
    title        = {{iGibson 2.0}: {O}bject-Centric Simulation for Robot Learning of Everyday Household Tasks},
    booktitle    = {CoRL 2021}
}

@inproceedings{roberts:2021,
    author       = {Mike Roberts AND Jason Ramapuram AND Anurag Ranjan AND Atulit Kumar AND Miguel Angel Bautista AND Nathan Paczan AND Russ Webb AND Joshua M. Susskind},
    title        = {Hypersim: {A} Photorealistic Synthetic Dataset for Holistic Indoor Scene Understanding},
    booktitle    = {ICCV 2021}
}

@inproceedings{shen:2021,
    author       = {Bokui Shen AND Fei Xia AND Chengshu Li AND Roberto Mart{\'i}n-Mart{\'i}n AND Linxi Fan AND Guanzhi Wang AND Claudia P{\'e}rez-D'Arpino AND Shyamal Buch AND Sanjana Srivastava AND Lyne P. Tchapmi AND Micael E. Tchapmi AND Kent Vainio AND Josiah Wong AND Li Fei-Fei AND Silvio Savarese},
    title        = {{iGibson 1.0}: {A} Simulation Environment for Interactive Tasks in Large Realistic Scenes},
    booktitle    = {IROS 2021}
}

@inproceedings{szot:2021,
    author       = {Andrew Szot AND Alex Clegg AND Eric Undersander AND Erik Wijmans AND Yili Zhao AND John Turner AND Noah Maestre AND Mustafa Mukadam AND Devendra Chaplot AND Oleksandr Maksymets AND Aaron Gokaslan AND Vladimir Vondrus AND Sameer Dharur AND Franziska Meier AND Wojciech Galuba AND Angel Chang AND Zsolt Kira AND Vladlen Koltun AND Jitendra Malik AND Manolis Savva AND Dhruv Batra},
    title        = {{Habitat 2.0}: {T}raining Home Assistants to Rearrange their Habitat},
    booktitle    = {NeurIPS 2021}
}

@inproceedings{fan:2022,
    author       = {Linxi Fan AND Guanzhi Wang AND Yunfan Jiang AND Ajay Mandlekar AND Yuncong Yang AND Haoyi Zhu AND Andrew Tang AND De-An Huang AND Yuke Zhu AND Anima Anandkumar},
    title        = {{MineDojo}: {B}uilding Open-Ended Embodied Agents with Internet-Scale Knowledge},
    booktitle    = {NeurIPS 2022 Datasets and Benchmarks Track}
}

@inproceedings{li:2022,
    author       = {Chengshu Li AND Ruohan Zhang AND Josiah Wong AND Cem Gokmen AND Sanjana Srivastava AND Roberto Mart{\'i}n-Mart{\'i}n AND Chen Wang AND Gabrael Levine AND Wensi Ai AND Benjamin Martinez AND Hang Yin AND Michael Lingelbach AND Minjune Hwang AND Ayano Hiranaka AND Sujay Garlanka AND Arman Aydin AND Sharon Lee AND Jiankai Sun AND Mona Anvari AND Manasi Sharma AND Dhruva Bansal AND Samuel Hunter AND Kyu-Young Kim AND Alan Lou AND Caleb R. Matthews AND Ivan Villa-Renteria AND Jerry Huayang Tang AND Claire Tang AND Fei Xia AND Yunzhu Li AND Silvio Savarese AND Hyowon Gweon AND C. Karen Liu AND Jiajun Wu AND Li Fei-Fei},
    title        = {{BEHAVIOR-1K}: {A} Human-Centered, Embodied {AI} Benchmark with 1,000 Everyday Activities and Realistic Simulation},
    booktitle    = {CoRL 2022}
}

@inproceedings{martinezgonzalez:2021,
    author       = {Pablo Martinez-Gonzalez AND Sergiu Oprea AND John A. Castro-Vargas AND Alberto Garcia-Garcia AND Sergio Orts-Escolano AND Jose Garcia-Rodriguez AND Markus Vincze},
    title        = {{UnrealROX+}: {A}n Improved Tool for Acquiring Synthetic Data from Virtual {3D} Environments},
    booktitle    = {IJCNN 2021}
}

@inproceedings{wright:2022,
    author       = {Daniel Wright AND Krzysztof Narkowicz AND Patrick Kelly},
    title        = {Lumen: {R}eal-time Global Illumination in {U}nreal {E}ngine 5},
    booktitle    = {SIGGRAPH 2022 Advances in Real-Time Rendering in Games}
}

@inproceedings{bordes:2023,
    author    = {Florian Bordes AND Shashank Shekhar AND Mark Ibrahim AND Diane Bouchacourt AND Pascal Vincent AND Ari S. Morcos},
    title     = {{PUG}: {P}hotorealistic and Semantically Controllable Synthetic Data for Representation Learning},
    booktitle = {NeurIPS 2023}
}

@inproceedings{gong:2023,
    author       = {Ran Gong AND Jiangyong Huang AND Yizhou Zhao AND Haoran Geng AND Xiaofeng Gao AND Qingyang Wu AND Wensi Ai AND Ziheng Zhou AND Demetri Terzopoulos AND Song-Chun Zhu AND Baoxiong Jia AND Siyuan Huang},
    title        = {ARNOLD: {A} Benchmark for Language-Grounded Task Learning With Continuous States in Realistic {3D} Scenes},
    booktitle    = {ICCV 2023}
}

@inproceedings{kirillov:2023,
    author       = {Alexander Kirillov AND Eric Mintun AND Nikhila Ravi AND Hanzi Mao AND Chloe Rolland AND Laura Gustafson AND Tete Xiao AND Spencer Whitehead AND Alexander C. Berg AND Wan-Yen Lo AND Piotr Doll{\'a}r AND Ross Girshick},
    title        = {Segment Anything},
    booktitle    = {ICCV 2023}
}

@inproceedings{langmead:2025,
    author       = {Arran Langmead},
    title        = {Procedural Content Generation in {UE5}},
    booktitle    = {Game Developers Conference 2023}
}

@inproceedings{mizrahi:2023,
    author       = {David Mizrahi AND Roman Bachmann AND O{\u{g}}uzhan Fatih Kar AND Teresa Yeo AND Mingfei Gao AND Afshin Dehghan AND Amir Zamir},
    title        = {{4M}: {M}assively Multimodal Masked Modeling},
    booktitle    = {NeurIPS 2023}
}

@inproceedings{bachmann:2024,
    author       = {Roman Bachmann AND O{\u{g}}uzhan Fatih Kar AND David Mizrahi AND Ali Garjani AND Mingfei Gao AND David Griffiths AND Jiaming Hu AND Afshin Dehghan AND Amir Zamir},
    title        = {{4M-21}: {A}n Any-to-Any Vision Model for Tens of Tasks and Modalities},
    booktitle    = {NeurIPS 2024}
}

@inproceedings{ge:2024,
    author       = {Yunhao Ge AND Yihe Tang AND Jiashu Xu AND Cem Gokmen AND Chengshu Li AND Wensi Ai AND Benjamin Jose Martinez AND Arman Aydin AND Mona Anvari AND Ayush K Chakravarthy AND Hong-Xing Yu AND Josiah Wong AND Sanjana Srivastava AND Sharon Lee AND Shengxin Zha AND Laurent Itti AND Yunzhu Li AND Roberto Martin-Martin AND Miao Liu AND Pengchuan Zhang AND Ruohan Zhang AND Li Fei-Fei AND Jiajun Wu},
    title        = {{BEHAVIOR} Vision Suite: {C}ustomizable Dataset Generation via Simulation},
    booktitle    = {CVPR 2024}
}

@inproceedings{ke:2024,
    author       = {Bingxin Ke AND Anton Obukhov AND Shengyu Huang AND Nando Metzger AND Rodrigo Caye Daudt AND Konrad Schindler},
    title        = {Repurposing Diffusion-Based Image Generators for Monocular Depth Estimation},
    booktitle    = {CVPR 2024}
}

@inproceedings{mazur:2024,
    author       = {Kirill Mazur AND Gwangbin Bae AND Andrew J. Davison},
    title        = {{SuperPrimitive}: {S}cene Reconstruction at a Primitive Level},
    booktitle    = {CVPR 2024}
}

@inproceedings{puig:2024,
    author       = {Xavi Puig AND Eric Undersander AND Andrew Szot AND Mikael Dallaire Cote AND Tsung-Yen Yang AND Ruslan Partsey AND Ruta Desai AND Alexander William Clegg AND Michal Hlavac AND So Yeon Min AND Vladim{\'i}r Vondru{\v{s}} AND Theophile Gervet AND Vincent-Pierre Berges AND John M. Turner AND Oleksandr Maksymets AND Zsolt Kira AND Mrinal Kalakrishnan AND Jitendra Malik AND Devendra Singh Chaplot AND Unnat Jain AND Dhruv Batra AND Akshara Rai AND Roozbeh Mottaghi},
    title        = {{Habitat 3.0}: {A} Co-Habitat for Humans, Avatars and Robots},
    booktitle    = {ICLR 2024}
}

@inproceedings{tong:2024,
    author       = {Shengbang Tong AND Ellis Brown AND Penghao Wu AND Sanghyun Woo AND Manoj Middepogu AND Sai Charitha Akula AND Jihan Yang AND Shusheng Yang AND Adithya Iyer AND Xichen Pan AND Ziteng Wang AND Rob Fergus AND Yann LeCun AND Saining Xie},
    title        = {{Cambrian-1}: {A} Fully Open, Vision-Centric Exploration of Multimodal {LLMs}},
    booktitle    = {NeurIPS 2024}
}

@inproceedings{bai:2025,
    author       = {Jianhong Bai AND Menghan Xia AND Xiao Fu AND Xintao Wang AND Lianrui Mu AND Jinwen Cao AND Zuozhu Liu AND Haoji Hu AND Xiang Bai AND Pengfei Wan AND Di Zhang},
    title        = {{ReCamMaster}: {C}amera-Controlled Generative Rendering from a Single Video},
    booktitle    = {ICCV 2025}
}

@inproceedings{narkowicz:2025,
    author       = {Krzysztof Narkowicz AND Tiago Costa},
    title        = {{MegaLights}: {S}tochastic Direct Lighting in {U}nreal {E}ngine 5},
    booktitle    = {SIGGRAPH 2025 Advances in Real-Time Rendering in Games}
}

@inproceedings{ravi:2025,
    author       = {Nikhila Ravi AND Valentin Gabeur AND Yuan-Ting Hu AND Ronghang Hu AND Chaitanya Ryali AND Tengyu Ma AND Haitham Khedr AND Roman R{\"a}dle AND Chloe Rolland AND Laura Gustafson AND Eric Mintun AND Junting Pan AND Kalyan Vasudev Alwala AND Nicolas Carion AND Chao-Yuan Wu AND Ross Girshick AND Piotr Doll{\'a}r AND Christoph Feichtenhofer},
    title        = {{SAM 2}: {S}egment Anything in Images and Videos},
    booktitle    = {ICLR 2025}
}

@inproceedings{wang:2025,
    author       = {Jianyuan Wang AND Minghao Chen AND Nikita Karaev AND Andrea Vedaldi AND Christian Rupprecht AND David Novotny},
    title        = {{VGGT}: {V}isual Geometry Grounded Transformer},
    booktitle    = {CVPR 2025}
}

@inproceedings{wu:2025a,
    author       = {Wayne Wu AND Honglin He AND Jack He AND Yiran Wang AND Chenda Duan AND Zhizheng Liu AND Quanyi Li AND Bolei Zhou},
    title        = {{MetaUrban}: {A}n Embodied {AI} Simulation Platform for Urban Micromobility},
    booktitle    = {ICLR 2025}
}

@inproceedings{wu:2025b,
    author       = {Wayne Wu AND Honglin He AND Chaoyuan Zhang AND Jack He AND Seth Z. Zhao AND Ran Gong AND Quanyi Li AND Bolei Zhou},
    title        = {Towards Autonomous Micromobility through Scalable Urban Simulation},
    booktitle    = {CVPR 2025}
}

@inproceedings{ye:2025,
    author       = {Xiaokang Ye AND Jiawei Ren AND Yan Zhuang AND Xuhong He AND Yiming Liang AND Yiqing Yang AND Xianrui Zhong AND Mrinaal Dogra AND Eric Liu AND Kevin Benavente AND Rajiv Mandya Nagaraju AND Dhruv Vivek Sharma AND Ziqiao Ma AND Tianmin Shu AND Zhiting Hu AND Lianhui Qin},
    title        = {{SimWorld}: {A}n Open-ended Simulator for Agents in Physical and Social Worlds},
    booktitle    = {NeurIPS 2025}
}

@inproceedings{zhong:2025,
    author       = {Fangwei Zhong AND Kui Wu AND Churan Wang AND Hao Chen AND Hai Ci AND Zhoujun Li AND Yizhou Wang},
    title        = {{UnrealZoo}: {E}nriching Photo-realistic Virtual Worlds for Embodied {AI}},
    booktitle    = {ICCV 2025}
}

@inproceedings{wang:2026,
    author    = {Jianyuan Wang AND Minghao Chen AND Shangzhan Zhang AND Nikita Karaev AND Johannes Sch{\"o}nberger AND Patrick Labatut AND Piotr Bojanowski AND David Novotny AND Andrea Vedaldi AND Christian Rupprecht},
    title     = {{VGGT-$\Omega$}},
    booktitle = {CVPR 2026}
}

@misc{vanrossum:2005,
    author       = {Guido van Rossum AND Alyssa Coghlan},
    title        = {{PEP 343}: {T}he ``with'' Statement},
    howpublished = {Python Enhancement Proposals}
}

@misc{beattie:2016,
    author       = {Charles Beattie AND Joel Z. Leibo AND Denis Teplyashin AND Tom Ward AND Marcus Wainwright AND Heinrich K{\"u}ttler AND Andrew Lefrancq AND Simon Green AND V{\'i}ctor Vald{\'e}s AND Amir Sadik AND Julian Schrittwieser AND Keith Anderson AND Sarah York AND Max Cant AND Adam Cain AND Adrian Bolton AND Stephen Gaffney AND Helen King AND Demis Hassabis AND Shane Legg AND Stig Petersen},
    title        = {{DeepMind Lab}},
    howpublished = {arXiv 2016}
}

@misc{brockman:2016,
    author       = {Greg Brockman AND Vicki Cheung AND Ludwig Pettersson AND Jonas Schneider AND John Schulman AND Jie Tang AND Wojciech Zaremba},
    title        = {{OpenAI Gym}},
    howpublished = {arXiv 2016}
}

@misc{kolve:2017,
    author       = {Eric Kolve AND Roozbeh Mottaghi AND Winson Han AND Eli VanderBilt AND Luca Weihs AND Alvaro Herrasti AND Daniel Gordon AND Yuke Zhu AND Abhinav Gupta AND Ali Farhadi},
    title        = {{AI2-THOR}: {A}n Interactive {3D} Environment for Visual {AI}},
    howpublished = {arXiv 2017}
}

@misc{savva:2017,
    author       = {Manolis Savva AND Angel X. Chang AND Alexey Dosovitskiy AND Thomas Funkhouser AND Vladlen Koltun},
    title        = {{MINOS}: {M}ultimodal Indoor Simulator for Navigation in Complex Environments},
    howpublished = {arXiv 2017}
}

@misc{szelei:2017,
    author       = {Tamás Szelei},
    title        = {rpclib: Modern msgpack-rpc for {C++}},
    howpublished = {\url{http://rpclib.net}}
}

@misc{vinyals:2017,
    author       = {Oriol Vinyals AND Timo Ewalds AND Sergey Bartunov AND Petko Georgiev AND Alexander Sasha Vezhnevets AND Michelle Yeo AND Alireza Makhzani AND Heinrich K{\"u}ttler AND John Agapiou AND Julian Schrittwieser AND John Quan AND Stephen Gaffney AND Stig Petersen AND Karen Simonyan AND Tom Schaul AND Hado van Hasselt AND David Silver AND Timothy Lillicrap AND Kevin Calderone AND Paul Keet AND Anthony Brunasso AND David Lawrence AND Anders Ekermo AND Jacob Repp AND Rodney Tsing},
    title        = {{StarCraft II}: {A} New Challenge for Reinforcement Learning},
    howpublished = {arXiv 2017}
}

@misc{juliani:2018,
    author       = {Arthur Juliani AND Vincent-Pierre Berges AND Ervin Teng AND Andrew Cohen AND Jonathan Harper AND Chris Elion AND Chris Goy AND Yuan Gao AND Hunter Henry AND Marwan Mattar AND Danny Lange},
    title        = {Unity: {A} General Platform for Intelligent Agents},
    howpublished = {arXiv 2018}
}

@misc{openai:2019:dota,
    author       = {{OpenAI} AND Christopher Berner AND Greg Brockman AND Brooke Chan AND Vicki Cheung AND Przemys{\l}aw D{\k e}biak AND Christy Dennison AND David Farhi AND Quirin Fischer AND Shariq Hashme AND Chris Hesse AND Rafa{\l} J{\'o}zefowicz AND Scott Gray AND Catherine Olsson AND Jakub Pachocki AND Michael Petrov AND Henrique P. d. O. Pinto AND Jonathan Raiman AND Tim Salimans AND Jeremy Schlatter AND Jonas Schneider AND Szymon Sidor AND Ilya Sutskever AND Jie Tang AND Filip Wolski AND Susan Zhang},
    title        = {Dota {2} with Large Scale Deep Reinforcement Learning},
    howpublished = {arXiv 2019}
}

@misc{openai:2019:rubiks,
    author       = {{OpenAI} AND Ilge Akkaya AND Marcin Andrychowicz AND Maciek Chociej AND Mateusz Litwin AND Bob McGrew AND Arthur Petron AND Alex Paino AND Matthias Plappert AND Glenn Powell AND Raphael Ribas AND Jonas Schneider AND Nikolas Tezak AND Jerry Tworek AND Peter Welinder AND Lilian Weng AND Qiming Yuan AND Wojciech Zaremba AND Lei Zhang},
    title        = {Solving {Rubik's Cube} with a Robot Hand},
    howpublished = {arXiv 2019}
}

@misc{jakob:2022,
    author       = {Wenzel Jakob},
    title        = {nanobind: tiny and efficient {C++/Python} bindings},
    howpublished = {\url{https://github.com/wjakob/nanobind}}
}

@misc{epic:2025,
    title        = {{Clair Obscur} takes home 9 {Game Awards} including {Game of the Year}},
    howpublished = {Epic Games Store News},
    year         = {2025}
}

@misc{obedkov:2025,
    title        = {{Black Myth}: {W}ukong Tops 25 Million Copies Sold as its Merchandise Sales Skyrocket in {C}hina},
    howpublished = {Game World Observer},
    year         = {2025}
}

@misc{simeoni:2025,
    author       = {Oriane Sim{\'e}oni AND Huy V. Vo AND Maximilian Seitzer AND Federico Baldassarre AND Maxime Oquab AND Cijo Jose AND Vasil Khalidov AND Marc Szafraniec AND Seungeun Yi AND Micha{\"e}l Ramamonjisoa AND Francisco Massa AND Daniel Haziza AND Luca Wehrstedt AND Jianyuan Wang AND Timoth{\'e}e Darcet AND Th{\'e}o Moutakanni AND Leonel Sentana AND Claire Roberts AND Andrea Vedaldi AND Jamie Tolan AND John Brandt AND Camille Couprie AND Julien Mairal AND Herv{\'e} J{\'e}gou AND Patrick Labatut AND Piotr Bojanowski},
    title        = {{DINOv3}},
    howpublished = {arXiv 2025}
}

@misc{yang:2025,
    author       = {Shusheng Yang AND Jihan Yang AND Pinzhi Huang AND Ellis Brown AND Zihao Yang AND Yue Yu AND Shengbang Tong AND Zihan Zheng AND Yifan Xu AND Muhan Wang AND Daohan Lu AND Rob Fergus AND Yann LeCun AND Li Fei-Fei AND Saining Xie},
    title        = {{Cambrian-S}: {T}owards Spatial Supersensing in Video},
    howpublished = {arXiv 2025}
}

@misc{gabeur:2026,
    author       = {Valentin Gabeur AND Shangbang Long AND Songyou Peng AND Paul Voigtlaender AND Shuyang Sun AND Yanan Bao AND Karen Truong AND Zhicheng Wang AND Wenlei Zhou AND Jonathan T. Barron AND Kyle Genova AND Nithish Kannen AND Sherry Ben AND Yandong Li AND Mandy Guo AND Suhas Yogin AND Yiming Gu AND Huizhong Chen AND Oliver Wang AND Saining Xie AND Howard Zhou AND Kaiming He AND Thomas Funkhouser AND Jean-Baptiste Alayrac AND Radu Soricut},
    title        = {Image Generators are Generalist Vision Learners},
    howpublished = {arXiv 2026}
}

@misc{magne:2026,
    author       = {Lo{\"\i}c Magne AND Anas Awadalla AND Guanzhi Wang AND Yinzhen Xu AND Joshua Belofsky AND Fengyuan Hu AND Joohwan Kim AND Ludwig Schmidt AND Georgia Gkioxari AND Jan Kautz AND Yisong Yue AND Yejin Choi AND Yuke Zhu AND Linxi "Jim" Fan},
    title        = {{NitroGen}: {A}n Open Foundation Model for Generalist Gaming Agents},
    howpublished = {arXiv 2026}
}

@misc{nvidia:2026,
    title        = {{NVIDIA Isaac Sim}},
    howpublished = {\url{https://developer.nvidia.com/isaac/sim}},
}

@misc{panda3d:2026,
    title        = {{Panda3D}},
    howpublished = {\url{https://www.panda3d.org}},
}

@misc{unity:2026,
    title        = {{Unity}},
    howpublished = {\url{https://unity.com}},
}

@misc{unrealengine:2026,
    title        = {{Unreal Engine}},
    howpublished = {\url{https://www.unrealengine.com}}
}

@misc{unrealengine:2026:blueprints,
    title        = {Blueprints Visual Scripting},
    howpublished = {Epic Games Developer Documentation}
}

@misc{unrealengine:2026:components,
    title        = {An Overview of Components},
    howpublished = {Epic Games Developer Documentation}
}

@misc{unrealengine:2026:console,
    title        = {Console Variables and Commands},
    howpublished = {Epic Games Developer Documentation}
}

@misc{unrealengine:2026:python,
    title        = {Scripting the {Unreal Editor} Using {Python}},
    howpublished = {Epic Games Developer Documentation}
}

@misc{unrealengine:2026:reflection,
    title        = {{Unreal Engine} Reflection System},
    howpublished = {Epic Games Developer Documentation}
}
